\documentclass[10pt,twocolumn,letterpaper]{article}

\usepackage[pagenumbers]{cvpr} 

\usepackage[dvipsnames, table]{xcolor}

\usepackage{graphicx}
\usepackage{booktabs}
\usepackage{float}
\usepackage{adjustbox}      
\usepackage{bm}             
\usepackage{multirow}       
\usepackage{booktabs}       
\usepackage[symbol]{footmisc}   
\usepackage[accsupp]{axessibility}  

\usepackage{tabularx}
\definecolor{tabfirst}{rgb}{1, 0.7, 0.7} 
\definecolor{tabsecond}{rgb}{1, 0.85, 0.7} 
\definecolor{tabthird}{rgb}{1, 1, 0.7} 

\definecolor{cvprblue}{rgb}{0.21,0.49,0.74}
\usepackage[pagebackref,breaklinks,colorlinks,citecolor=cvprblue]{hyperref}

\def\paperID{297} 
\def\confName{3DV\xspace}
\def\confYear{2025\xspace}

\title{6Img-to-3D: Few-Image Large-Scale Outdoor Novel View Synthesis}

\author{%
  Théo Gieruc$^{1,2,*}$ 
  \quad
  Marius Kästingschäfer$^{1,3,*}$ 
  \quad
  Sebastian Bernhard$^{1}$
  \quad
  Mathieu Salzmann$^{2}$ \\
  $^1$Continental
  \quad
  $^2$SDSC \& CVLab, EPFL 
  \quad
  $^3$University of Freiburg \\
  \texttt{marius.kaestingschaefer@continental.com}
}

\begin{document}

\twocolumn[{%
\renewcommand\twocolumn[1][]{#1}%
\maketitle
\begin{center}
    \centering
    \captionsetup{type=figure}
    \includegraphics[width=1\textwidth]{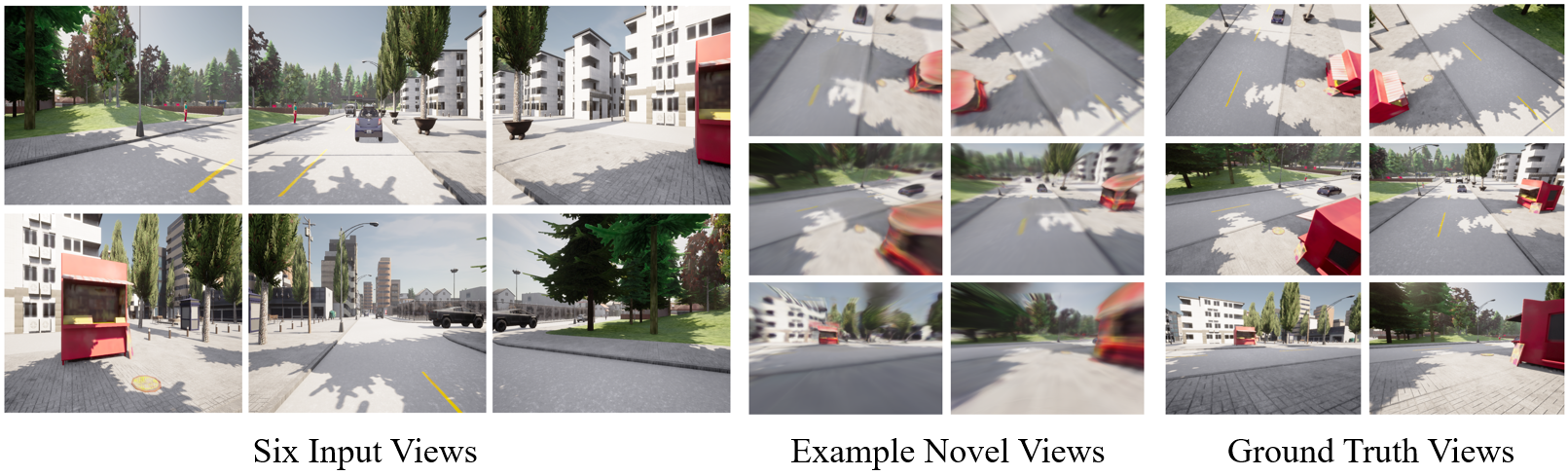}
    \captionof{figure}{\textbf{Qualitative results of 6Img-to-3D.} During inference time, 6Img-to-3D takes six surround vehicle RGB images with a small overlap as input and returns a parameterized triplane from which arbitrary novel views can be rendered.}
    \label{fig:introduction_results_overview}
\end{center}%
}]

\begin{abstract}
\vspace{-0.15cm}
Current 3D reconstruction techniques struggle to infer unbounded scenes from a few images faithfully. Most existing methods have high computational demands, require detailed pose information, and cannot reconstruct occluded regions reliably. We introduce 6Img-to-3D, a novel transformer-based encoder-renderer method for single-shot image-to-3D reconstruction. Our method outputs a 3D-consistent parameterized triplane from only six outward-facing input images for large-scale, unbounded outdoor driving scenarios. We take a step towards resolving existing shortcomings by combining contracted custom cross- and self-attention mechanisms for triplane parameterization, differentiable volume rendering, scene contraction, and image feature projection. We showcase on synthetic data that six surround-view vehicle images from a single timestamp are enough to reconstruct 360$^{\circ}$ scenes during inference time, taking 395 ms. Our method allows, for example, rendering third-person images and birds-eye views. Code, and more results are available at our \href{https://6Img-to-3D.GitHub.io/}{project page}.
\vspace{-0.25cm}



\end{abstract}

\section{Introduction}
\label{sec:intro}

Inferring the appearance and geometry of large-scale outdoor scenes from a few camera inputs is a challenging, unsolved problem. The problem is characterized by the complexity inherent in outdoor scenes, namely a vast spatial extent, diverse object textures, and ambiguity of the 3D geometry due to occlusions. Robotics and autonomous driving systems both require methods that process vision-centric first-person views from a fixed set of cameras as input and derive suitable control commands from these \cite{probabilistic_robotics_book_2005}. The ability to instantaneously perform 2D-to-3D scene reconstructions would be useful within such methods and thus have broad applicability within robotics \cite{adamkiewicz2022visiononly, ortiz2022isdf, wu2022daydreamer} and autonomous driving \cite{chen2023endtoend, zheng2023occworld, bogdoll2023muvo} domains. Quickly generating unobstructed bird's-eye views of challenging parking scenarios or of large vehicles for teleoperation could significantly aid human drivers.
For applying one-shot images to high-fidelity 3D techniques in vehicles or robots, additional factors, such as the inference speed, the quality of the inferred representation, the scalability of the approach, and the number of camera parameters required for the technique to work, are equally important. For those safety-critical domains, avoiding overly compressing the 3D scene or altering scene details by adding or omitting road obstacles is crucial.

\def\thefootnote{*}\footnotetext{These authors contributed equally to this work.}

\begin{figure}[ht!]
    \centering
    \includegraphics[width=1\linewidth]{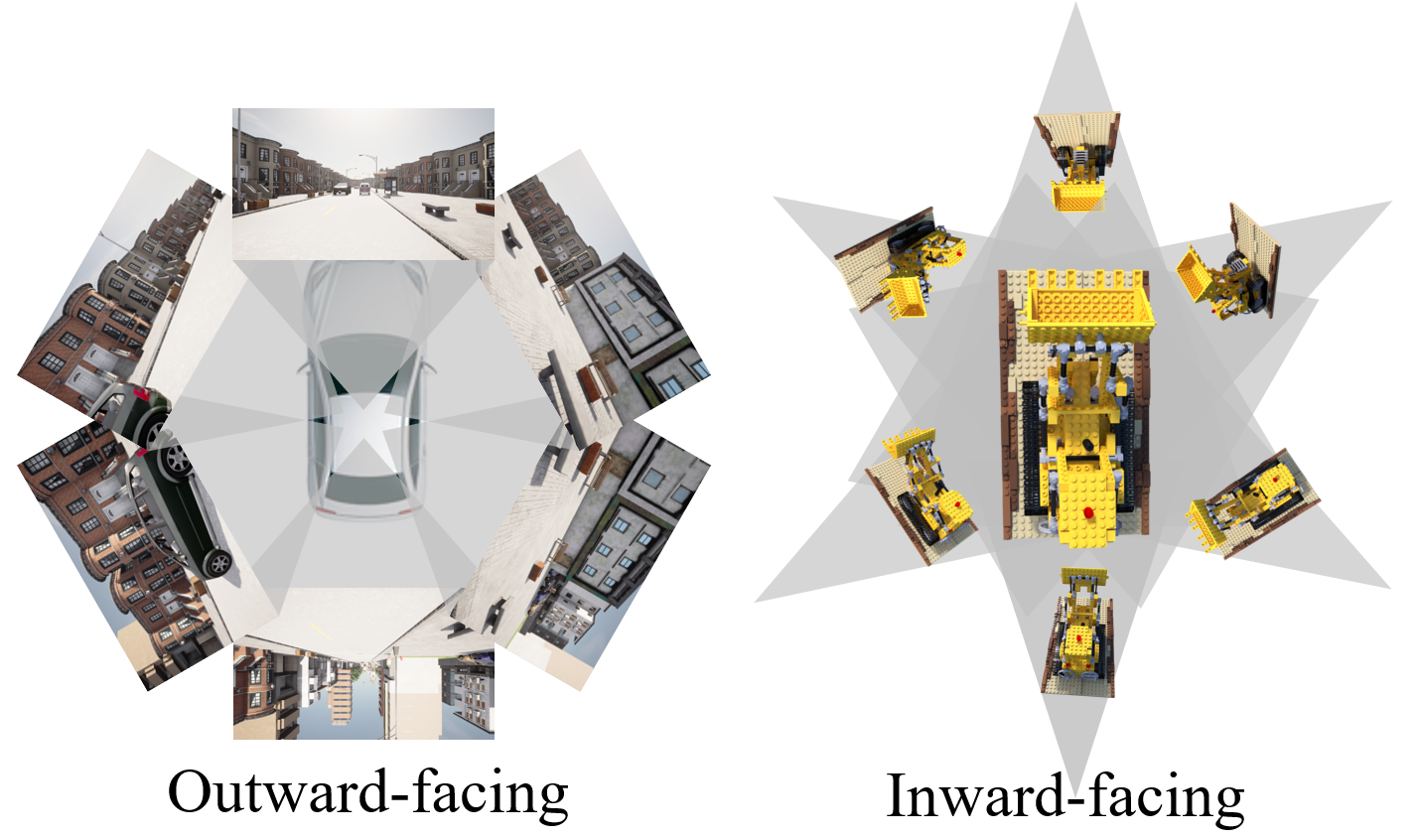}
    \caption{\textbf{View Overlap.} Inward and outward-facing camera setups differ significantly in their view overlap. Ourward-facing (inside-out) camera setups overlap minimally, whereas inward-facing (outside-in) setups can overlap across multiple cameras. The source of the excavator images is \cite{synthetic_NeRF_dataset, mildenhall_nerf_2020}.}
    \label{fig:view_overlap}
\end{figure}

Despite recent progress in 3D reconstruction from a few or single 2D images, many methods are either limited to single objects \cite{hong2023lrm, szymanowicz2023splatter, liu2023zero1to3, liu2023one2345, xu2023dmv3d} or to generative indoor view synthesis \cite{charatan2023pixelsplat}. Existing few-shot methods developed for single objects are often not easily transferable to large-scale or unbounded outdoor settings. This is because outdoor settings are more diverse due to their varying structural composition, occlusions, and vast extent. Another limitation is that many methods require a high degree of overlap between the input images, such as found in spherically arranged and inward-facing camera setups focused on a single object in the center \cite{mildenhall_nerf_2020, yu2021pixelnerf}. This is opposed to the outward-facing camera setup essential for autonomous driving where the camera overlap is minimal \cite{behley_semantickitti_2019, caesar_nuscenes_2020, Sun_2020_CVPR}. The camera setups are contrasted in \cref{fig:view_overlap}. Traditional methods in autonomous driving have shown promising results with such camera setups but lack visual fidelity since they only predict semantic occupancy \cite{huang2023triperspective, sima2023scene, wei2023surroundocc} and commonly rely on additional sensors such as LiDAR.

\vspace{-0.15cm}

\textbf{Problem Statement.} Our work addresses the challenge of designing, training, and evaluating a novel single-shot model for novel view synthesis able to render birds-eye and third-person perspectives of unbounded traffic scenes solely from six outward-facing ego vehicle images with minimal overlap. The relative camera poses are fixed and known. For training and evaluation, a synthetic dataset is used due to the absence of a suited real-world dataset.

Given the above limitations and the problem statement, we present \textbf{6Img-to-3D}, a transformer-based decoder-renderer method for single-shot image-to-3D reconstruction. 6Img-to-3D is fully differentiable and end-to-end trained on a single GPU. It simultaneously takes six outward-facing images (\textbf{6Img-}) as inputs and generates a parameterized triplane (\textbf{to-3D}) from which novel scene views can be rendered. Unlike previous methods, we do not focus on single objects but instead on complex driving scenarios characterized by a high-depth complexity recorded only by sparse multi-view outward-facing RGB cameras. Furthermore, unlike other methods, we do not use a multi- or few-shot scheme where an initial reconstruction is tested and improved iteratively. Instead, we output the 3D scene in a single shot without iterative model refinement. To this end, we utilize pre-trained ResNet features, cross- and self-attention mechanisms for parameterizing the triplane, and image feature projection to condition the renderer. Our method is trained using roughly 11.4K input and 190K supervising images from 1900 scenes. We do not require depth or LiDAR information. Due to the absence of suitable datasets, including a large number of non-ego vehicle views needed during training, we have to rely on synthetic data from Carla \cite{dosovitskiy2017carla}. \newline

\noindent The contributions of this paper can be summarized as follows: 

\begin{enumerate}
    \item \textbf{Novel Architecture.} We present 6Img-to-3D, a novel single-shot, few-image novel view synthesis pipeline for large-scale outdoor environments. Our method outputs a parameterized triplane of the 3D scene from which arbitrary viewpoints can be rendered. We combine triplane-based differentiable volume rendering, renderer conditioning on projected image features, self- and custom cross-attention mechanisms, and an LPIPS loss. 
    \item  \textbf{Visual Fidelity.} We demonstrate that our method is, unlike other approaches, able to faithfully reconstruct unobserved scene parts. Our method outperforms existing few-image methods such as SplatterImage both in terms of depth and visual fidelity. 
    \item \textbf{Testing and Abblations.} We perform extensive additional tests and ablation studies to confirm the effect of individual model components such as the LPIPS loss, projected image features conditioning, and scene contraction.
    \item \textbf{Hardware Efficiency.} Unlike other large-scale 2D-to-3D methods, training 6Img-to-3D requires a single 42GB GPU and for inference only 10.5GB. From six-image-to-triplane our architecture only takes 395ms.
\end{enumerate}


\section{Related Work}
\label{sec:background}

\noindent \textbf{3D Representations.} 
Traditional methods for representing 3D scenes include voxels \cite{VoxelColoring1997, girdhar2016learning, Xie_2020, wu2017marrnet}, point clouds \cite{fan2016point, yu2021pointr, yang2018foldingnet, achlioptas2018learning, zeng2022lion}, signed distance field (SDF) \cite{wang2021neus} and polygon meshes \cite{kanazawa2018learning, wang2018pixel2mesh, nash2020polygen, xu2021disn, liu2023meshdiffusion}. Recently, numerous new implicit and explicit representations have emerged for learning 3D scenes \cite{Tewari2022NeuRendSTAR, xie2022neural, gao2023nerf, kerbl3Dgaussians}. 
Implicit neural fields such as Neural Radiance Fields (NeRFs) model the surrounding appearance and geometry using a continuously queriable function updated through differentiable volumetric rendering \cite{niemeyer2020differentiable, mildenhall_nerf_2020, Lombardi_2019, tagliasacchi2022volume, gao2023nerf}, either utilize a single \cite{chen2019learning, mildenhall_nerf_2020, zhang_nerf_2020} or multiple \cite{reiser_kilonerf_2021, rebain_derf_2020} multilayer perceptrons (MLPs). However, these models are slow to train, data-intensive, and expensive to render since querying the implicit neural field is computationally demanding.
Explicit approaches such as Plenoctrees \cite{yu_plenoctrees_2021}, Plenoxels \cite{yu_plenoxels_2021}, and NSVF \cite{liu2020neural} employ trilinear interpolation within a 3D grid to represent the scene. This significantly improves rendering and optimization time but comes at the expense of limited scalability due to the curse of dimensionality when increasing the resolution or scene size. 3D Gaussian Splats \cite{kerbl3Dgaussians} as another explicit approach do not face those downsides since they do not require empty space to be explicitly modeled.
Hybrid models \cite{sun_direct_2022, muller_instant_2022} adopt a voxel-like grid as an intermediary representation. Some models further refine this process by decomposing the voxel grid into three or more orthogonal planes called triplanes or K-Planes \cite{chen_tensorf_2022, chan_efficient_2022, bautista_gaudi_2022, fridovich-keil_k-planes_2023}. This enables the computation of features for each point in space by aggregating the feature values of projected 3D points onto each plane, transitioning from an N-cube voxel grid to a 3 $\times$ N-squared plane representation.

\noindent \textbf{Few-Image to 3D Representation.} Single-scene 3D reconstruction methods require many images to generalize towards novel views since the reconstruction problem would be underconstrained otherwise \cite{mildenhall_nerf_2020}. Hence, few-image to 3D representation methods incorporate additional global or local information to further regularize and constrain the problem.
Global regularization methods rely on additional model priors, either via applying further model regularization \cite{yang2023freenerf, niemeyer_regnerf_2021} or by utilizing pre-trained image models to provide an extra guidance signal to inform the optimization of the single-scene  \cite{deng2022nerdi, raj2023dreambooth3d, liu2023one2345, melaskyriazi2023realfusion, seo2023let, xu2023neurallift360, xu2023dmv3d, lin2023consistent123, po2023state}.
Local regularization methods train a cross-scene multi-perspective aggregator. Such models include PixelNeRF \cite{yu2021pixelnerf}, IBRNet \cite{wang_ibrnet_2021}, MVSNeRF \cite{chen2021mvsnerf}, VolRecon \cite{ren_volrecon_2023} and others \cite{trevithick2021grf, liu2022neural, johari2022geonerf, lin_vision_2022}. They retrieve image features via view projection and aggregate resulting features to obtain novel views.
We also apply pixel feature conditioning, but unlike our method, many of the mentioned local regularization methods rely on a large overlap between input images ensured by using spherical inward-facing cameras. Another category of methods does not train the representation on a per-scene basis but instead trains a meta-network that outputs the discrete parameterized scene representation. Models such as \cite{shue_3d_2022, gao_get3d_2022, gu_nerfdiff_2023, anciukevicius_renderdiffusion_2023, watson_novel_2022} use diffusion to generate 3D representations of scenes. Many of those models use the triplane representation \cite{shue_3d_2022, gao_get3d_2022, gu_nerfdiff_2023, anciukevicius_renderdiffusion_2023, bhattarai_triplanenet_2023}, whereas some directly generate in the image-space  \cite{watson_novel_2022} or can be conditioned on input images \cite{gu_nerfdiff_2023, watson_novel_2022, anciukevicius_renderdiffusion_2023}. 
The recent single-shot method SplatterImage \cite{szymanowicz23splatter} parameterizes several pixel-aligned Gaussians as scene representation. While being fast to render, this method struggles with occlusions and unobserved scene parts. The method that most closely resembles our work is LRM \cite{hong2023lrm}, which utilizes cross- and self-attention to parameterize a triplane using a few object images. However, LRM does not handle unbound scenes, is trained on 128 NVIDIA A100, and has seven times more parameters than our method.

\noindent \textbf{Large-Scale 3D Scene Representations.} Extending NeRFs to unbounded scenes requires spatially contracting the scene, which is usually done proportionally to disparity as introduced in MipNeRF-360 \cite{barron_mip-nerf_2022} or using an inverted sphere parametrization as in NeRF++ \cite{zhang2020nerf}. Architectures modeling outdoor driving scenes are BlockNeRF \cite{tancik_block-nerf_2022}, DisCoScene \cite{xu2022discoscene}, NeuralField-LDM \cite{kim2023neuralfieldldm} and Neural ground planes \cite{sharma2023neural}. Those are, however, not dealing with sparse input views.
Neo360 \cite{irshad_neo_2023} utilizes local image features to infer an image-conditional triplanar representation, whereby the model dissociates foreground from background scene parts. Unlike our method, Neo360 is trained on 8 A100 GPUs and focuses on inward-facing camera perspectives with large view overlap as input to the model. Large-scale scene generation methods such as Infinite Nature \cite{liu2021infinite, li2022infinitenaturezero} and others often produce additional images in an autoregressive fashion without the ability to input camera coordinates for novel view synthesis control \cite{koh2021pathdreamer, xie2023citydreamer, lin2023infinicity, Chen_2023, chai2023persistent}. In autonomous driving, established methods create a bird's-eye-view (BEV) semantic representation of a scene using either Inverse Perspective Mapping (IPM) \cite{reiher_sim2real_2020} or attention mechanisms \cite{yang_projecting_2021, li_bevformer_2022}. TPVFormer \cite{huang_tri-perspective_2023}, a work that our architecture is partially based on, maps six outward-facing vehicle 2D images onto three orthogonal planes using self- and cross-deformable attention mechanisms to obtain semantic occupancy predictions. The method is, however, supervised with depth information from a LiDAR sensor and, unlike our method, only predicts semantic occupancy.
\section{6Img-to-3D}
\label{sec:methods}

\begin{figure*}[t!]
    \centering
    \includegraphics[width=1\textwidth]{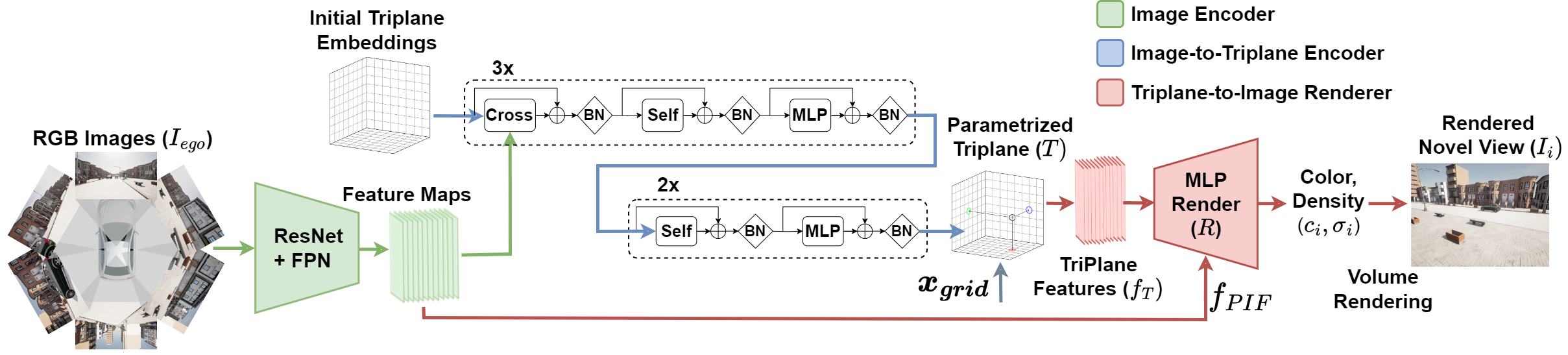}
    \caption{\textbf{The overview of 6Img-to-3D.} Given six input images, we first encode them into feature maps using a pre-trained ResNet and an FPN (\cref{sec:image_encoder}). The scene coordinates are contracted to fit the unbounded scenes (\cref{sec:triplane_representation}).  MLPs, cross-and self-attention layers form the Image-to-Triplane Encoder of our framework (\cref{sec:image_triplane_encoder}). Images can be rendered from the resulting triplane using our renderer (\cref{sec:triplane_image_encoder}). We additionally condition the rendering process on projected image features.}
    \label{fig:method_6Img-to-3D}
\end{figure*}

Given six outward-facing images $\bm{I}_{ego}$, their camera extrinsic (relative pose) matrices $\bm{M}$ and associated camera intrinsic matrices $\bm{K}$ with appropriate dimensions, the goal is to reconstruct the surrounding 3D occupancy and appearance, that is the scene $\bm{S}$. Since the values of the scene $\bm{S}$ are not known, we use ${N}$ spherical multi-view images $\bm{I}_{sphere}$ with their associated extrinsic and extrinsic matrices during training as a supervision signal to estimate the scene $\bm{\hat{S}}$. The proposed architecture for lifting $\bm{I}_{ego}$ consists of an image-to-triplane encoder. After a forward pass through this architecture, the discrete triplane contains the approximation of the scene $\bm{\hat{S}}$ from which novel views $\bm{I}_{i}$ can be rendered. The following sections describe the different parts of the pipeline visualized in \cref{fig:method_6Img-to-3D} and the training objectives.

\subsection{Image Encoder} \label{sec:image_encoder}
Six RGB input images $\bm{I}_{ego} \in \mathbb{R}^{6 \times 3 \times H \times W} $ are processed via a pre-trained ResNet \cite{he2015deep} followed by a Feature Pyramid Network (FPN) \cite{lin2017feature}, resulting in multi-scale pixel-aligned image features  $\bm{F}_I \in \mathbb{R}^{6 \times 3 \times H/i \times W/i},$ with $i=\{8,16,32,64\}$.

\subsection{Triplane Representation} \label{sec:triplane_representation}
The triplane consists of three pairwise orthogonal feature grids ${\bm{T}_{HW}}$, ${\bm{T}_{HZ}}$ and ${\bm{T}_{WZ}}$, each representing parts of the decomposed 3D space, a setup similar to EG3D \cite{chan_efficient_2022}. We choose triplanes because they allow gradient-based optimization, are explicit representations and their parameterization is thus straightforward. Additionally, they are lightweight and provide an easily queriable information encoding \cite{chan_efficient_2022, fridovich-keil_k-planes_2023}. The triplane ${\bm{T}}$ is composed of the three coordinate-centered and axis-aligned planes ${\bm{T}_{HW}}$, ${\bm{T}_{HZ}}$ and ${\bm{T}_{WZ}}$, respectively of size $200 \times 200 \times F_{T}$, $200 \times 16 \times F_{T}$ and $200 \times 16 \times F_{T}$, where $F_{T}$ is the number of feature channels of the triplane. Due to the concentration of relevant information near the ground plane in 3D outdoor driving scenes, we decide to allocate more space to the horizontal dimensions ($HW$) than to the vertical dimension ($Z$). The grid coordinates $\bm{x_{grid}}$ for each plane are normalized between -1 and 1. The contraction equation from world coordinates $\bm{x_w}$ to grid coordinates $\bm{x_{grid}}$ is given as element-wise operations by
\begin{align}\label{eq:contr}
\begin{split}
    \bm{x_{Ws}} &= \bm{x_W}\cdot \bm{s}\\
    \bm{x_{grid}} &= \begin{cases}
\text{${\frac{\bm{x_{Ws}}}{2}} $} & \text{$\|\bm{x_{Ws}}\| \leq 1$}\\
\text{$(2-\frac{1}{\|\bm{x_{Ws}}\|}) \frac{\bm{x_{Ws}}}{2\|\bm{x_{Ws}}\|}$} & \text{$\|\bm{x_{Ws}}\| > 1$}
\end{cases}
\end{split}
\end{align}
In short, a scaling parameter $\bm{s} = [s_h,s_w,s_z]$ is applied before contraction to adapt to the nature of our scenes, and the scaled world coordinates $\bm{x_{Ws}}$ are contracted using the spatial distortion introduced in Mip-NeRF 360 \cite{barron_mip-nerf_2022} to obtain $\bm{x_{grid}}$, with $||\cdot||$  denoting the L2 norm. 

\subsection{Image-to-Triplane Encoder} \label{sec:image_triplane_encoder}
The Image-to-Triplane Encoder, adopted from TPVFormer \cite{huang2023triperspective}, comprises deformable custom cross- and self-attention layers, both using residual connections. The first half of the encoder consists of three consecutive blocks of cross-attention, self-attention, and MLPs, and the second half consists of two blocks that only have self-attention layers and MLPs. The next sections will introduce each component in more detail. Intermediate batch normalization (BN) layers are applied within both blocks. The number of blocks is empirically motivated. We hypothesize that increasing the number of blocks could further improve performance; the current design reflects a balance chosen considering our computing constraints. LRM \cite{hong2023lrm}, for example, uses 16 blocks in total. \\

\noindent \textbf{Cross-Attention (Image-Triplane)} We use cross-attention to incorporate as much information as possible from input image features into the triplane grids. This is facilitated by deformable attention mechanisms (DeformAttn) \cite{zhu_deformable_2021}. Unlike traditional attention mechanisms, this layer is designed to handle the high dimensionality of both the image features and the triplanar grids by restricting the attention of each query to a small set of keys, rather than to all available keys as in traditional attention. This layer is applied to each plane separately. Establishing the link between each query (triplane feature) and its keys (image features) is non-trivial. We compute the deformable attention the following way: 
\begin{equation*}
\begin{split}
    \textbf{CrossAttn}(\bm{T_{k}}) = \text{DeformAttn}(\bm{T_{k}}, \bm{p^{F}_{k}}, \bm{F_{I}}), \\
    k \in \{HW, HZ, WZ\}\ 
\end{split}
\end{equation*}

whereby the grid features $\bm{T_{k}}$ are used as queries, the image features $\bm{F_{I}}$ as keys and the reference points $\bm{p^{F}_{k}}$ links a selected number of keys to each query.

\begin{figure}[ht]
    \centering
    \includegraphics[width=1\linewidth]{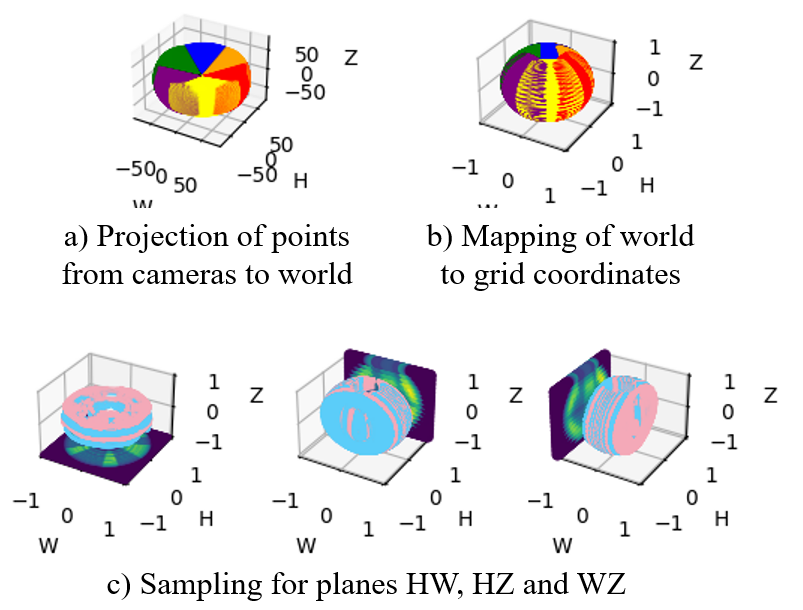}
    \caption{\textbf{Establishing Reference Points.} a) Points are projected into the world space from the cameras. b) Those points are mapped into the grid space. c) For each plane element, $n$ points are selected to be used as reference points between the grid index and image feature.}
    \label{fig:cross-attention}
\end{figure}

We establish the correspondence via reference points $\bm{p^{F}_{k}}$, computed as follows: we first establish a correspondence from camera indices $(u,v)_{cam_i}$ to world coordinates $\bm{x}_{\text{W}}$ by sampling points along each input camera ${cam_i}$ ray, as illustrated in \cref{fig:cross-attention}a. The world coordinates $\bm{x}_{\text{W}}$ are then transformed into grid coordinates $\bm{x}_{\text{grid}}$, employing \cref{eq:contr}, as shown in \cref{fig:cross-attention}b. Given the obtained correspondences between camera indices $(u,v)_{cam_i}$ and 3D grid coordinates, for each plane $\bm{T_{k}}$ we consider only a subset of $n_k$ fixed slices of the correspondences to obtain the reference points $p_F^k$, illustrated in \cref{fig:cross-attention}c. For any $(i,j)$ index in the plane $\bm{T_{k}}$, we now have a correspondence to $n_k$ image feature indexes via the reference points $\bm{p^{F}_{k}}$. $n_k$ is proportional to the dimension perpendicular to the plane $\bm{T_{k}}$ ($4$ for ${\bm{T}_{HW}}$, $32$ for ${\bm{T}_{HZ}}$ and ${\bm{T}_{WZ}}$). 

\noindent \textbf{Self-Attention (Triplane)} This layer is the same as the one within TPVFormer. It allows the three planes to exchange information between and within themselves. The self-attention layer also makes use of the deformable attention described above: 
\begin{equation*}
\begin{split}
\textbf{SelfAttn}(\bm{T_{k}}) = \text{DeformAttn}(\bm{T_{k}}, \bm{p^{T}_{k}}, \mathbf{T}), \\
k \in \{HW, HZ, WZ\}\;.
\end{split}
\end{equation*}

whereby $\bm{T_{k}}$ is the plane used as query, the triplane $\mathbf{T}$ is used as keys and the query to keys correspondence is defined by the reference points $\bm{p^T_{k}}$. For any element in a plane, reference points are obtained by randomly sampling in their neighborhood and in their perpendicular direction. 

\subsection{Triplane-to-Image Renderer} \label{sec:triplane_image_encoder}

The renderer $\bm{R}$, consisting of a shallow MLP, predicts the color $\bm{c}$ and density $\sigma \in [0,1]$ for a sampled 3D point $\bm{x_W} \in \mathbb{R}^3$. As we are dealing with Lambertian scenes, the view directions are discarded. The sampling point $\bm{x_W}$ is first converted into grid coordinates $\bm{x_{grid}}$ using \cref{eq:contr}. The triplane features $\bm{f_{T}}$ are then obtained by projecting this point via bilinear sampling onto each plane and applying the Hadamard product (elementwise multiplication) on those plane features: $\bm{f_{T}}^{ijk} =\bm{T}_{HW}(\bm{x_{grid}}^{ij}) \cdot \bm{T}_{HZ}(\bm{x_{grid}}^{ik}) \cdot \bm{T}_{WZ}(\bm{x_{grid}}^{jk})$. 

Inspired by PixelNeRF \cite{yu_pixelnerf_2021}, we extract the Projected Image Features $\bm{f_{PIF}}$ from the pixel-aligned input image features $\bm{F_I}$ by projecting $\bm{x_W}$ back into the image space of the input cameras: ${(u,v)}_{cam_i} = \bm{K}_i ~ \bm{M}_i^{-1} ~ {\bm{x_{W}}}, i \in \{1,...,6\}$. Since each point in space can be seen by a maximum of two cameras, the aggregation function concatenates the valid features, with zero-padding when one or no camera sees the point: $\bm{f_{PIF}} = \text{aggregation}(F_{Ii}\text{(u,v)}_{cam_i} \text{ for } i \in \{1,...,6\})$.

As the triplane features $\bm{f_{T}}$ and the projected image features $\bm{f_{PIF}}$ come from two different distributions, we first process $\bm{f_{PIF}}$ through batch normalization before going through the renderer: $\bm{c}, \sigma = \bm{R}(\bm{f_{T}}, \textrm{BatchNorm}(\bm{f_{PIF}})) $. Given the volume density $\sigma_i$ and the color $\bm{c_i}$, we follow the same volume rendering equations as NeRF \cite{mildenhall_nerf_2020}, originating from \cite{volume_rendering_1995}. Initially, multiple coarse points are sampled uniformly along each ray $\bm{r}$. Those coarse points are then fed into the renderer to obtain densities, which are used for finer proportional sampling. The renderer then accumulates contributions from these points along the ray, which can be expressed as
\begin{equation}
    \hat{C}(\bm{r})=\sum_{i=1}^N T_i\left(1-\exp \left(-\sigma_i \delta_i\right)\right) \mathbf{c}_i\;,
\end{equation}
where $\bm{r}$ represents a ray with origin $\bm{o}$ and direction $\bm{d}$, $T_i=\exp \left(-\sum_{j=1}^{i-1} \sigma_j \delta_j\right)$ is the ray transmission upto sample $i$, $1-\exp \left(-\sigma_i \delta_i\right)$ is the absorption from sample $i$, and $\delta_i=t_{i+1}-t_i$ is the distance between adjacent samples. Hereby $t_i$ is the accumulated distance from the ray origin to the start of the $i$th segment, and $\hat{C}(\bm{r})$ is the accumulated RGB color, resulting in a pixel value corresponding to the origin $\bm{o}$ of the ray $\bm{r}$. We can compute each pixel value using the volume rendering equation to obtain a novel view $\bm{I}_i$.

\subsection{Training Objectives} 

Training is conducted using MSE and VGG-based ${L}_{\text{LPIPS}}$ losses on randomly sampled views, as well as the TV loss introduced by K-Planes on the triplanes $\bm{T}$ \cite{fridovich-keil_k-planes_2023} and a distortion loss from Mip-NeRF 360 \cite{barron_mip-nerf_2022}. This yields the objective
\begin{equation}
\begin{split}
    \mathcal{L} = \mathcal{L}_{color}( \bm{C}(\bm{r}), \hat{\bm{C}}(\bm{r}) ) +   \lambda_{t}\mathcal{L}_{TV}(\bm{T})   \\
     + \lambda_{d}\mathcal{L}_{dist}(\bm{s}, \bm{w}) +
     \lambda_{l}\mathcal{L}_{\text{LPIPS}}(\bm{x}, \hat{\bm{x}})\;.
\end{split}
 \end{equation}

The color loss is a simple MSE loss computed per ray over batches of size $N$. It can be written as
\begin{equation}
    \mathcal{L}_{color}( \bm{C}(\bm{r}), \hat{\bm{C}}(\bm{r}) ) = \frac{1}{N} \sum_{i=1}^{N} ( \bm{C}(\bm{r}_i), \hat{\bm{C}}(\bm{r}_i) )^{2}\;.
\end{equation}

\noindent Similar to \cite{fridovich-keil_k-planes_2023, yu_plenoxels_2021, chen_tensorf_2022}, the total variation in space regularization is applied to smoothen the triplanes. This is done via the loss
\vspace{0.1cm}
\begin{equation}
    \mathcal{L}_{TV}(\bm{T}) = \frac{1}{ 3n^{2}} \sum_{k,i,j} \left( \|\bm{T}_{k}^{i,j}-\bm{T}_{k}^{i-1,j}\|_{2}^{2} + \|\bm{T}_{k}^{i,j}-\bm{T}_{k}^{i,j-1}\|_{2}^{2} \right)
\end{equation}
\vspace{0.1cm}
for $\bm{T_{k}}$ in $\{\bm{T}_{HW}, \bm{T}_{HZ}, \bm{T}_{WZ}\}$, where $i, j$ are the grid indices of the plane. To reduce floating artifacts and enforce volume rendering weights to be compact, we regularize the scene using  MipNeRF360's \cite{barron_mip-nerf_2022} distortion loss, given by
\vspace{0.1cm}
\begin{equation}
\begin{split}
    \mathcal{L}_{dist}(\bm{s}, \bm{w}) = \sum_{i, j} w_i w_j \bigg| \frac{s_i + s_{i+1}}{2} - \frac{s_j + s_{j+q}}{2} \bigg| \\
    + \frac{1}{3} \sum_i w_{i}^{2} (s_{i+1} - s_i)\;,
\end{split}
\end{equation}
\vspace{0.1cm}
where $\bm{s}$ is a set of ray distances and $\bm{w}$ are the volume rendering weights parameterizing each ray, following \cite{barron_mip-nerf_2022}.
\section{Experiments}
\label{sec:results}

\subsection{Dataset}
We use a synthetic dataset containing both ego and non-ego vehicle views few-view image novel view synthesis in an autonomous driving setting. Concerning novel view synthesis, other approaches also rely on synthetic data \cite{irshad_neo_2023}. The dataset contains 2000 single-timestep complex outdoor driving scenes, each offering six outward-facing vehicle images and 100 spherical images for supervision. The dataset also consists of five multi-timestep scenes, each with 200 consecutive driving steps containing six outward-facing vehicle cameras, one bird's-eye view, and one camera following the vehicle from behind. We only use the multi-timestep data for visualization. This results in 220K individual RGB images, each with their associated intrinsic camera matrix and pose. The dataset represents various driving scenes, vehicle types, pedestrians, and lighting conditions. To show the generalizability of our method, we use CARLA \cite{dosovitskiy2017carla} Town 1, Town 3 to 7, and 10 for generating training data, resulting in 1900 training scenes, and left the 100 scenes from Town 2 for testing. For further details, we refer to the publication of the dataset \cite{SEED4D_unpublished}. 

\subsection{Implementation Details}

\textbf{Model Inputs.} The six input images have a resolution of $1600 \times 928$ each, and the supervision images are downscaled to $64 \times 48$ pixels. \\
\textbf{Architecture.} Our triplanes have feature channels $F_T$ of size $128$. The renderer $R$ is implemented as a five-layer MLP with 128 neurons per layer, taking as input $128$ triplane features $f_{T}$ and $2 \times 128$ projected image features $f_{PIF}$. \\
\textbf{Training.} All models were trained on a single Nvidia A40 GPU with 42GB of VRAM for $100$ epochs, resulting in a training time of five days, with an Adam optimizer \cite{kingma2017adam}, a learning rate of $5\mathrm{e}\text{-}5$, and a cosine scheduler \cite{loshchilov2017sgdr} with 1000 warmup steps. One epoch consists of 1900 steps, each comprising a new scene and three randomly sampled views as supervision, scaled to $64 \times 48$ pixels. Using more than three supervising views per scene leads to diminishing returns; using only three views thus reduces the total training time, a phenomenon also observed by others \cite{hong2023lrm}. Ray-based sampling allows our model trained on low-resolution images to render high-resolution outputs during inference. We sample 64 coarse followed by 64 fine points during training. \\
\textbf{Inference.} A forward pass to obtain the parameterized triplane is completed in 395 ms. Rendering an image of size $400 \times 300$ from this triplane, using 128 uniformly-spaced points along each ray, takes 520 ms without the Projected Image Features (PIFs) vs 1461 ms when using PIFs. To improve the quality, a second pass based on the density of the first uniform sampling can be done by resampling 128 points in denser regions. This brings the rendering speed to 955ms without PIFs and 2853 ms with PIFs. Adding or removing the PIFs and sampling in a single or double fashion thus offers a trade-off between visual fidelity and speed. While being trained on a 40GB GPU, our method requires as little as 10.5 GB of vRAM during inference.

\subsection{Quantitative Results}

\begin{figure*}[t]
    \centering
    \includegraphics[width=1\textwidth]{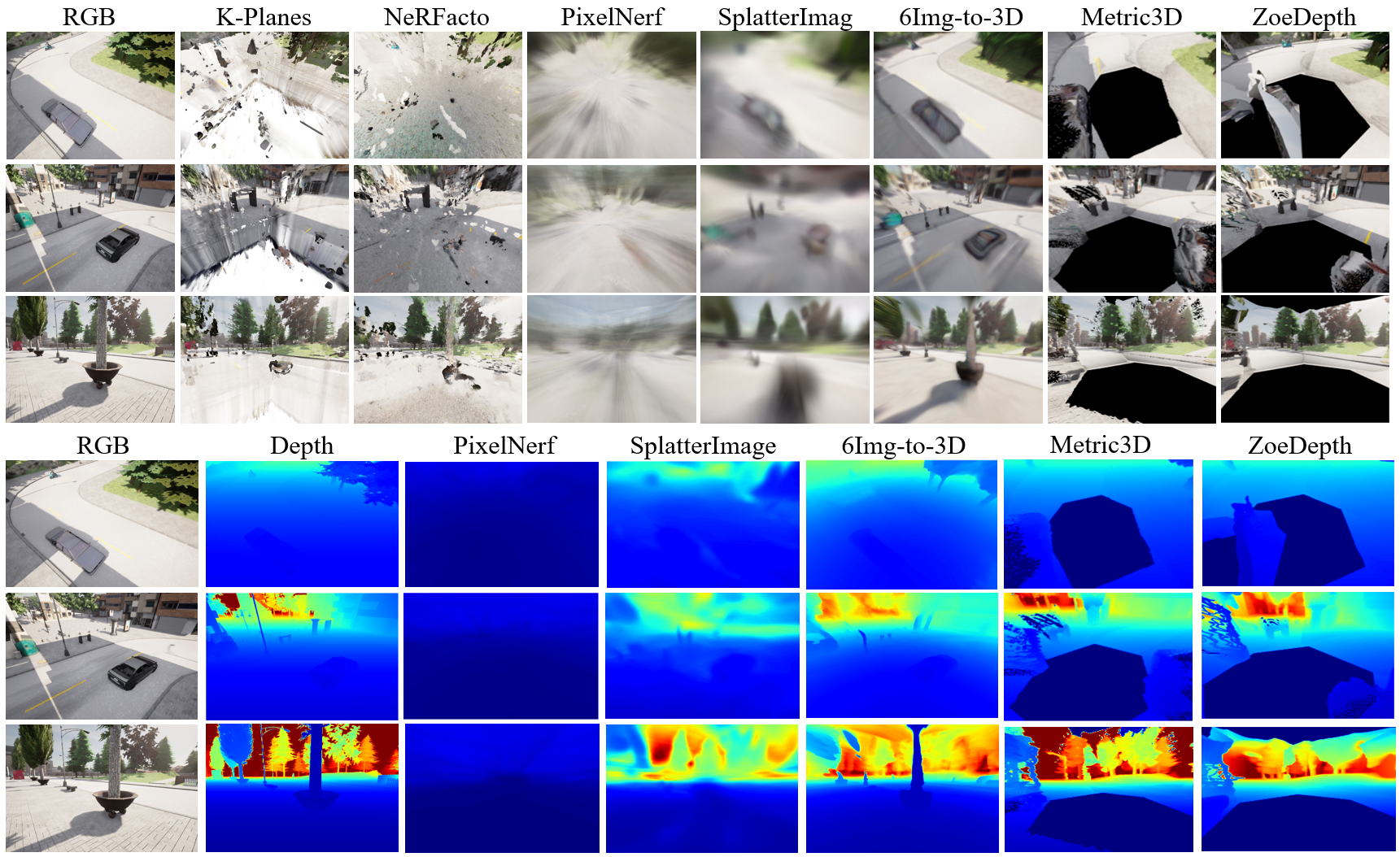}
    \caption{\textbf{Qualitative Results.} PixelNeRF and SplatterImage visibly struggle to model details and cannot resemble texture properly. Our method represents shape and appearance faithfully. 6Img-to-3D infers a full 3D representation of the scene and can thus also be queried for density only, resulting in reasonable depth maps. Due to their low performance, we do not visualize SplatFacto at the top and SplatFacto K-Planes and NeRFacto at the bottom part.}
    \label{fig:comparison_visualization}
\end{figure*}

We evaluate our method (6Img-to-3D) against several approaches such as SplatFacto  \cite{ye2023mathematical}, K-Planes \cite{fridovich-keil_k-planes_2023}, NeRFacto \cite{nerfstudio_Tancik_2023}, the related few-view methods PixelNeRF \cite{yu2021pixelnerf} and the recent SplatterImage \cite{szymanowicz23splatter}. We also compare against two metric monocular depth estimation techniques, namely ZoeDepth \cite{bhat2023zoedepth} and Metric3D \cite{hu2024metric3d}.
\begin{table}[!h]
    \centering
    \caption{\textbf{Result Comparison.} Due to occlusion artifacts we also compute all metrics for ZoeDepth and Metric3D while masking out regions occluded without points, indicated with a ‡. For a fair comparison, the unmasked values are considered when ranking the methods. Colors indicate \colorbox{tabfirst}{first}, \colorbox{tabsecond}{second}, and \colorbox{tabthird}{third} best values.}
    \vspace{-0.2cm}
    \label{tab:results_3D}
    \setlength{\tabcolsep}{2pt}
    \begin{tabular}{l c c c c}
        \toprule
          Methods & \textbf{PSNR} $\uparrow$ & \textbf{SSIM}$\uparrow$ & \textbf{LPIPS}$\downarrow$ & \textbf{DRMSE}$\downarrow$ \\
         \midrule
          ZoeDepth \cite{bhat2023zoedepth} & 5.466  &  0.254 &   \cellcolor{tabthird}0.563 &  11.728 \\
          ZoeDepth$^{\text{‡}}$  \cite{bhat2023zoedepth} & 14.202  &  0.661 &   0.292 &  9.378 \\
          Metric3D \cite{hu2024metric3d} & 6.314  & 0.296 &  \cellcolor{tabsecond}0.554 &  \cellcolor{tabsecond}10.049 \\
          Metric3D$^{\text{‡}}$ \cite{hu2024metric3d} & 13.699 &  0.600 &  0.336 &  8.655 \\
          NeRFacto \cite{nerfstudio_Tancik_2023}  & 10.943  &  0.298 &  0.791 & -- \\
          K-Planes \cite{fridovich-keil_k-planes_2023} & 11.356 &  0.463 &  0.633 & -- \\
          SplatFacto \cite{ye2023mathematical} & 11.607  &  0.486 &   0.658 &  -- \\
          PixelNeRF \cite{yu_pixelnerf_2021} & \cellcolor{tabthird}14.500 & \cellcolor{tabthird}0.550 & 0.652 & 19.235 \\
          SplatterImage \cite{szymanowicz23splatter}  & \cellcolor{tabsecond}17.791  & \cellcolor{tabsecond}0.580 &  0.568  & \cellcolor{tabthird}11.049 \\ 
          \bottomrule
          6Img-to-3D & \cellcolor{tabfirst}18.683 & \cellcolor{tabfirst}{0.726} & \cellcolor{tabfirst}{0.451} & \cellcolor{tabfirst}{6.232} \\         
        \bottomrule
    \end{tabular}
\end{table}


We would have liked to compare against Neo360's approach, however, the method requires a considerable computational budget for training, i.e., 8 A100 GPUs, which was not available for this paper.
%
Performance is measured using the established metrics PSNR, LPIPS \cite{zhang2018unreasonable}, and MSE on the test set. The results can be found in \cref{tab:results_3D}. We evaluated SplatFacto, K-Planes, and NeRFacto by training them on a subset of the evaluation set. Due to the negligible camera view overlap, the iterative methods cannot reconstruct the scene. PixelNeRF and SplatterImage are trained and evaluated under conditions equivalent to our method. To evaluate Metric3D and ZoeDepth, monocular depth maps are generated from ego images and then unprojected into space to create a colored point cloud, from which novel views can be rendered. For details of the evaluation protocols, see the appendix.
\vspace{0.05cm}

\subsection{Qualitative Results} \label{sec:qualitative_results}

\begin{figure*}[ht!]
    \centering
    \includegraphics[width=0.9\textwidth]{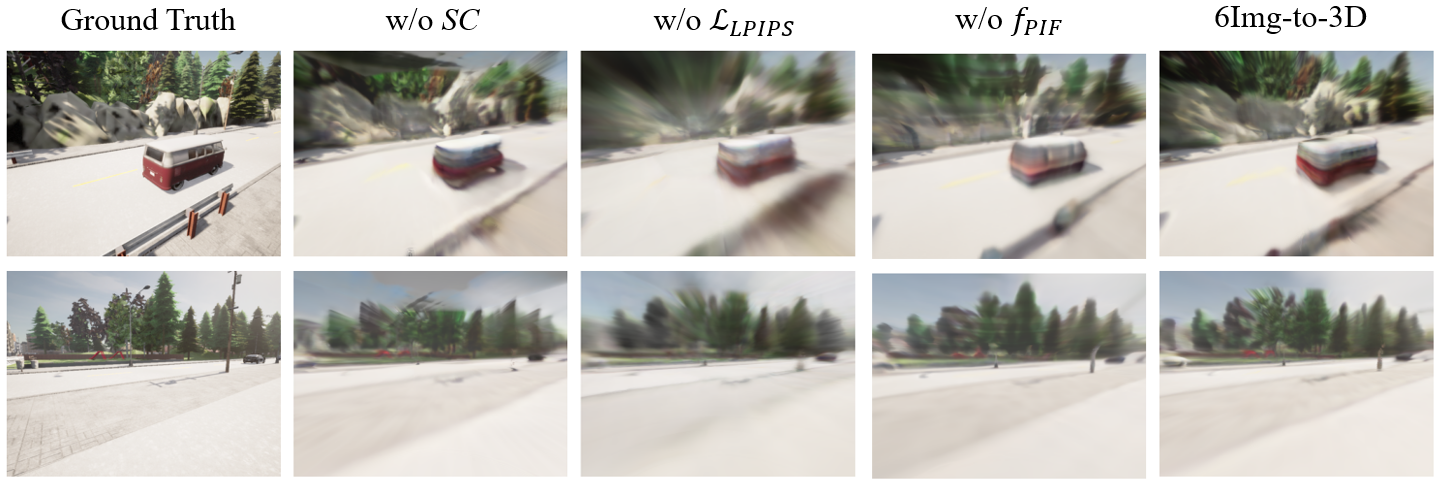}
    \caption{\textbf{Visualization of the ablation.} The final model particularly benefits from the LPIPS loss, the scene contraction, and the PIFs. When not using Scene Contraction, the out-of-bound objects are not rendered. Without the PIFs, the model struggles with fine details. Removing the LPIPS loss leads to increased smearing artifacts potentially caused by the projection of the PIFs along the camera rays.}
    \label{fig:qualitative_ablation}
\end{figure*}

\noindent \textbf{Novel View Visualization.} When comparing the visual results we find our method to be visually most appealing. Novel views resulting from projecting monocular depth values into space are often misaligned and unobserved parts lead to visible artifacts. PixelNerf and SplatterImage struggle to model texture and geometry faithfully. Since our method performs single-shot inference over occluded or unobserved parts of the scene those inferred parts are less sharp than in ground truth images. Results are visualized in the upper part of \cref{fig:comparison_visualization}.

\noindent \textbf{Depth Visualization.} We visualize the expected termination depth of each ray, resulting in the depth maps shown in the bottom part of \cref{fig:comparison_visualization}. For the colored point clouds resulting from the unprojected depth images, the closest point along the viewing direction per pixel is rendered. While never trained on depth information explicitly, 6Img-to-3D outputs reasonable depth maps for novel views. 

\subsection{Ablation Study}
We show the effectiveness of our model elements in \cref{tab:ablation_features_table} and visualize our ablations in \cref{fig:qualitative_ablation}. All model variants are trained for 100 epochs and evaluated on the holdout validation set consisting of 100 scenes from town two. We find that the scene contraction combined with the contracted cross-attention mechanism substantially improves the model performance. The PIFs and LPIPS help adjust the color values further. While LPIPS leads to more human-appealing images, this comes at the cost of reduced PSNR performance. Double and single sampling here refers to rendering the image with or without a more fine-grained second sampling step. Additionally, we performed ablation studies on how well our architecture scales with data, plotted in appendix \cref{sec:effects_scaling}.

\begin{table}[H]
    \caption{\textbf{Ablation Study.} The colors indicate \colorbox{tabfirst}{best}, \colorbox{tabsecond}{second}-best, and \colorbox{tabthird}{third}-best values.} 
    \label{tab:ablation_features_table}
    \centering
    \small
    \setlength{\tabcolsep}{2pt}
\begin{tabular}{lcccccc}
    \toprule
      \multicolumn{1}{l}{Method} &
      \multicolumn{2}{c}{\textbf{PSNR} $\uparrow$} &
      \multicolumn{2}{c}{\textbf{SSIM} $\uparrow$} &
      \multicolumn{2}{c}{\textbf{LPIPS} $\downarrow$} \\
    Sampling &
      \multicolumn{1}{l}{Single} &
      \multicolumn{1}{l}{Double} &
      \multicolumn{1}{l}{Single} &
      \multicolumn{1}{l}{Double} &
      \multicolumn{1}{l}{Single} &
      \multicolumn{1}{l}{Double} \\ 
      \midrule
    w/o \textit{SC}  & 17.155 & 17.352 & 0.699 & 0.721 & 0.487 & 0.474\\
    w/o $\mathcal{L}_{\text{LPIPS}}$ &    \cellcolor{tabfirst}18.674    & \cellcolor{tabfirst}18.812 &  \cellcolor{tabfirst}0.718     & \cellcolor{tabfirst}0.731 &   \cellcolor{tabthird}0.551    &  \cellcolor{tabthird}0.537 \\
    w/o $f_{PIF}$ & \cellcolor{tabthird}18.201 & \cellcolor{tabthird}18.256 &  \cellcolor{tabsecond}0.714 & \cellcolor{tabthird}0.718 & \cellcolor{tabsecond}0.482 & \cellcolor{tabsecond}0.487 \\
    \midrule
    6Img-to-3D               & \cellcolor{tabsecond}18.567 & \cellcolor{tabsecond}18.683 & \cellcolor{tabthird}0.712 & \cellcolor{tabsecond}0.726 & \cellcolor{tabfirst}0.451 & \cellcolor{tabfirst}0.451 \\ 
    \bottomrule
\end{tabular}
\end{table}

\vspace{0.05cm}

\section{Discussion}\label{sec:discussion}

\noindent \textbf{Conclusion.} In this paper, we present 6Img-to-3D, a novel, high-fidelity method for reconstructing 3D representations of unbounded outdoor scenes from a few images. Our evaluation demonstrates that 6Img-to-3D can faithfully reconstruct third-person vehicle perspectives and birds-eye views using only six images with minimal overlap and without depth information. We achieve this by leveraging a triplanar representation of space, deformable attention mechanisms, feature projection, and scene contraction. Our pipeline benefits from utilizing the features from the input images at multiple pipeline stages. A custom cross-attention sampling mechanism for contracted planes essentially drives performance gains. Unlike for example SplatterImage, our method is natively designed to incorporate multiple images at once and integrate their information into a single scene representation. We demonstrate the potential of our approach, as a single-shot high-fidelity novel view synthesis method for unbounded scenes. We consider this problem to be important for applications including the teleoperation of large vehicles (such as trucks) and vision-based parking assistance systems. 

\textbf{Future Directions.} Since we are using querying-based color and density computation our rendering is less fast than for example Gaussian Splatting. Future work includes porting our rendering pipeline towards Gaussian Splats.

We made some first attempts to run our trained model zero-shot on the nuScenes dataset \cite{caesar_nuscenes_2020}. Those tests show promising results that the model can zero-shot be extended to real-world data, for detail see appendix \cref{sec:zero_shot_nuScenes}. Using sim-to-real transfer methods could aid this process. We assume that real-world driving data could be utilized by leveraging multi-timestep images for training or fine-tuning whereby LiDAR could provide an additional depth supervision signal.

Further, tuning scene contraction and other hyperparameters used within our method could help reduce blurring artifacts. Similarly, additional data and longer training could improve performance. We conclude this from the absence of overfitting during training and from how the model scales with data. An increased model size allows for a higher triplane resolution, which might help alleviate blurry texture for fine-grained details. Our model has around $\frac{1}{7}$ the amount of parameters of LRM \cite{hong2023lrm}, so visual fidelity could certainly be improved by scaling the model. 

During initial experiments with an upsampler trained on the data for post-rendering upscaling improved PSNR, and SSIM scores but introduced blurring. See appendix \cref{sec:comp_annex} for further information. Incorporating the six source image information into the upscaling process represents a potential research direction to improve performance. \\


\noindent \textbf{Acknowledgements} The research leading to these results is partially funded by the German Federal Ministry for Economic Affairs and Climate Action within the project “NXT GEN AI METHODS". The authors wish to extend their sincere gratitude to the creators of TPVformer, NeRFStudio, and the KPlanes paper for generously open-sourcing their code.

\vspace{-0.05cm}
{
    \small
    \bibliographystyle{ieeenat_fullname}
    \bibliography{main}
}
\clearpage
\setcounter{page}{1}
\maketitlesupplementary

\section{Appendix}\label{sec:appendix}

\subsection{Bounded Scene Paramterization}

The transformation from world coordinates $\bm{x_{W}}$ to grid coordinates $\bm{x_{grid}}$ can also be done when a bounded scene is given.

\noindent\textbf{Bounded scenes:} The world coordinates $\bm{x_W}$ are transformed into grid coordinates $\bm{x_{grid}}$ by scaling the scene with the scaling parameter $\bm{s} = [s_h,s_w,s_z]$ and applying the offset $\bm{o} = [o_,o_w,o_z]$ according to \cref{eq:world2grid} for element-wise operations. When depth maps are available, out-of-bound pixels of the training images can be masked with a white background during an initial testing phase, so the network learns only to reconstruct the inside of the bounded scene.
\begin{align}
\label{eq:world2grid}
    \bm{x_{grid}} = \frac{\bm{x_W} - \bm{o}}{\bm{s}} 
\end{align}

\subsection{Shape and Parameter Details}
In our model description, we follow the PyTorch convention of Batch $\times$ Channel $\times$ Height $\times$ Width ($B  \times C \times H \times W$). The shape of 6Img-to-3D is shown in \cref{tab:model_parameter}. While the size of the input images during training remains fixed, the shape of the rendered images can be varied during both training and inference. 

\begin{table}[ht!]
    \centering
    \caption{Shape and parameter values of 6Img-to-3D trained on the introduced dataset.}
    \label{tab:model_parameter}
    \setlength{\tabcolsep}{2pt}
    \begin{tabular}{lccc}
        \toprule
        \textbf{Part}        & \textbf{Shape}                                                 & \textbf{Parameters} \\ 
        \midrule
        Input Images         & $6\times3\times928\times1600$                                                   & n.a.                \\
        \midrule
        ResNet Maps         & $6\times512\times116\times200$                                                  & 44M                 \\
                            & $6\times1024\times58\times100$                                                  &                     \\
                            & $6\times2048\times29\times50$                                                   &                     \\
        Feature Maps        & $6\times128\times116\times200$                                                  & 1M                  \\
                            & $6\times128\times58\times100$                                                   &                     \\
                            & $6\times128\times29\times50$                                                    &                     \\
                            & $6\times128\times15\times25$                                                    &                     \\
        Attention Mechanism &         n.a. & 16M                 \\
        Triplane            & $128\times200\times200$                                                    &            n.a.         \\
                            & $128\times200\times16$                                                     &                     \\
                            & $128\times16\times200$       &                     \\
        MLP decoder         &input: 384 & 84K                 \\
                            &output: 4 &                 \\
        \midrule
        Output Image        & $3\times600\times800$                    &  \\
        \bottomrule
    \end{tabular}
\end{table}

\subsection{Baselines and Competing Methods}

\subsubsection{Depth unprojection}

\begin{figure}[ht!]
    \centering
    \includegraphics[width=1\linewidth]{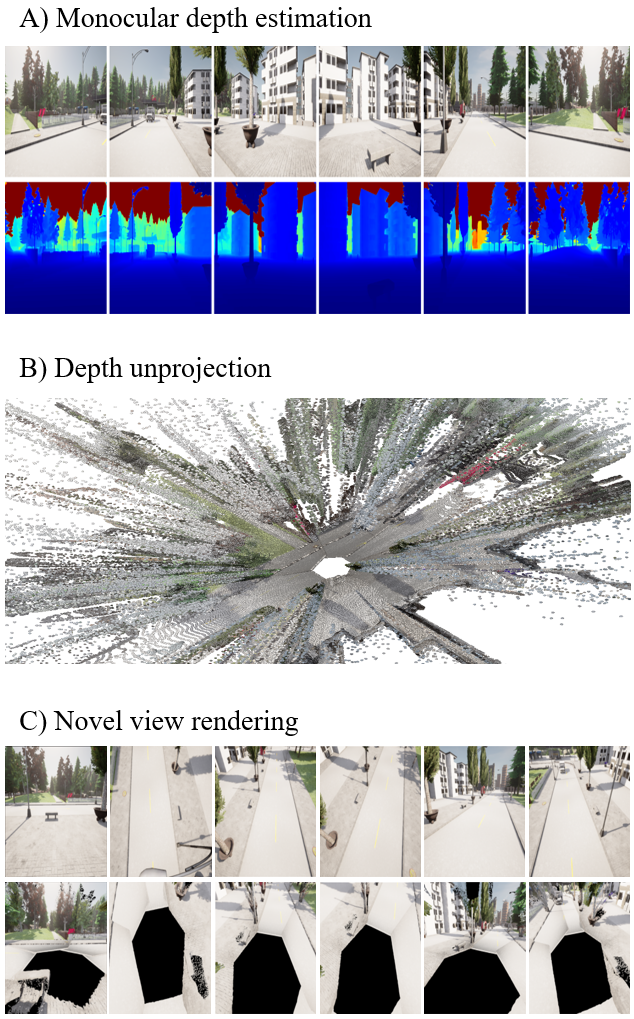}
    \caption{\textbf{Depth Unprojection.} A monocular depth method is used to obtain depth maps for the ego views (step A). Depth maps and RGB ego views are then used to obtain a colored point cloud (step B). From the colored point cloud novel views can be rendered (step C).}
    \label{fig:depth_unprojection}
\end{figure}

\noindent \textbf{ZoeDepth}~\cite{bhat2023zoedepth} is a zero-shot metric depth estimation technique. The method is trained on multiple datasets, including the autonomous driving dataset KITTI. The method builds on the MiDaS depth estimation framework. We utilize the model's publicly available code along with pre-trained weights that have been trained on both the NYU Depth v2 and KITTI datasets.
\noindent \textbf{Metric3D}~\cite{hu2024metric3d} is a metric 3D reconstruction method using a canonical camera space transformation method. The method can perform zero-shot metric depth and surface normal estimation from a single image. The publicly available code is used, and the largest version of the pre-trained model, vit\_giant, is used.

\noindent  \textbf{Unprojection steps:} Since our method reconstructs a Lambertian scene, we compare it against a depth baseline. We use one of the monocular depth estimation methods to obtain a depth map for each of the six ego input images resized to $842 \times 842$ to fit the model. Since camera intrinsics and extrinsic are known, we can use the depth maps to project the image pixels into space to obtain a colored point cloud (sometimes also referred to as 2.5D). The obtained colored point cloud can now be used to rasterize novel exo views. The pipeline is visualized in \cref{fig:depth_unprojection}. We optimized the available parameters (e.g., the size of the rendered points) to obtain the best possible results. We use open3d and pytorch3d to implement the pipeline.
\vspace{-0.2cm}

\subsubsection{SplatterImage}
\noindent \textbf{SplatterImage}~\cite{szymanowicz23splatter} is designed for inferring 3D Gaussian Splatting primitives from conditioning images in a pixel-aligned fashion. A U-net style image-to-primitive mapping network supported by a cross-attention mechanism maps the input RGB images to a 'Splatter Image' containing opacity, position, shape, and color information of a 3D Gaussian. We use the repository released by the authors.
\vspace{-0.2cm}

\subsubsection{PixelNeRF}
\noindent \textbf{PixelNeRF}~\cite{yu_pixelnerf_2021} is a sparse novel view synthesis method. PixelNeRF weakens some of the shortcomings of the original NeRF paper by leveraging projected image features and training across multiple scenes. We use the re-implementation introduced in the code of Neo360 \cite{irshad_neo_2023}.

\subsubsection{Iterative training methods}

\textbf{K-Planes}~\cite{fridovich-keil_k-planes_2023} factorizes a 3D scene into multiscale planes. The plane features are learned using differentiable volume rendering, sampled using multiscale bilinear interpolation, and rendered using a small MLP. We use the hybrid version of the model and the GitHub users Giodiro's reimplementation of the model available within NeRFstudio.

\noindent \textbf{NeRFacto}~\cite{nerfstudio_Tancik_2023} is a combination of several published methods. The method is optimized to work particularly well for real data captures. The following techniques are combined in this method: camera pose refinement, per-image appearance embedding learning, proposal sampling, scene contraction, and hash encoding. We use the model as part of NeRFStudio.

\noindent \textbf{SplatFacto}~\cite{kerbl20233d} is a re-implementation of the original 3D Gaussian Splatting paper~\cite{kerbl20233d} within NeRFStudio. The method explicitly stores a collection of 3D volumetric Gaussians to parameterize the scene. During rendering the 3D Gaussions are 'splatted' to obtain per-pixel colors. We use the model as part of NeRFStudio.

\subsection{Training Details}

\textbf{K-Planes.} We train each of the models for 30k steps on a single Tesla T4 GPU with 16GB of VRAM. We follow the model's default NeRFStudio \cite{nerfstudio_Tancik_2023} settings for training. Near and far bounds of the scene are adjusted to 0.1 to 60 to best accommodate the scenes. Additionally, scene contraction is applied. The training took around 1.5 hours per model.

\noindent \textbf{NeRFacto.} We train each of the models for 30k steps on a single Tesla T4 GPU with 16GB of VRAM. We follow the model's default NeRFStudio \cite{nerfstudio_Tancik_2023} settings for training. We disable the model's use of an appearance embedding since those lead to problems during the evaluation, and we also deactivate the camera pose optimization because we already provide the model with ground truth poses. The near and far bounds are set to 0.1 and 60. Each model is trained for a total of 1 hour.

\noindent \textbf{SplatFacto.} We train each of the models for 30k steps on a single Tesla T4 GPU with 16GB of VRAM. We again follow the model's default NeRFStudio \cite{nerfstudio_Tancik_2023} settings for training. The model took a total of 20 minutes to train.

\noindent \textbf{PixelNeRF.} We train PixelNeRF for 100k steps on a Nvidia A40 GPU with 42GB of VRAM, with an Adam optimizer \cite{kingma2017adam} and a learning rate of $1\mathrm{e}\text{-}3$. Total training time accumulates to five days.

\noindent \textbf{SplatterImage.} We train SplatterImage for five days across five 3090 GPUs with 24GB of VRAM. During training, the supervision images are scaled to $128 \times 128$ pixels. We use the multi-input image variant of the model to accommodate all six input views.

\noindent \textbf{ZoeDepth} and \textbf{Metric3D} are not fine-tuned.

\begin{figure}[ht!]
    \centering
    \includegraphics[width=1\linewidth]{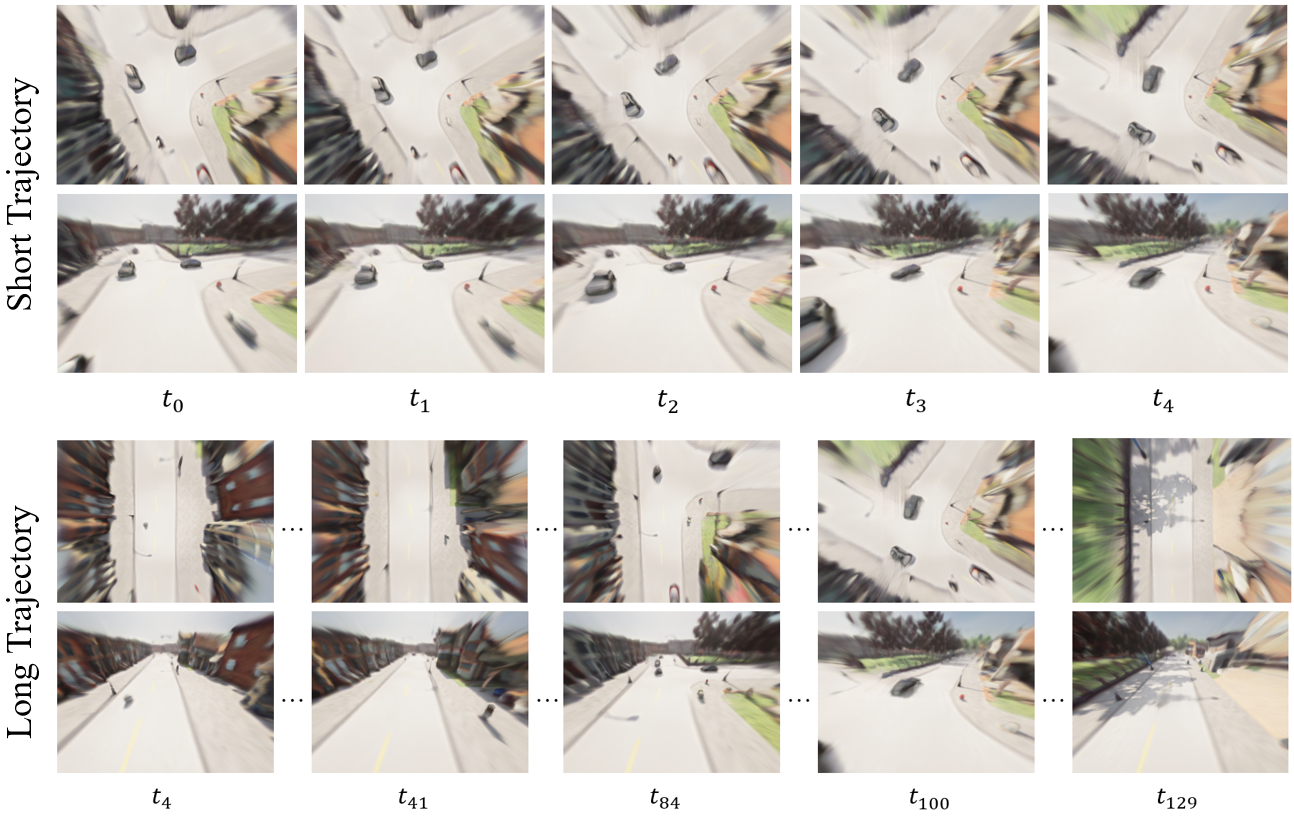}
    \caption{\textbf{Trajectory Visualization.} Consecutive timesteps are visualized with images from a BEV camera and a view location behind and above the ego vehicle.}
    \label{fig:trajectories}
\end{figure}

\subsection{Consistency Over Timesteps} 
Our model works deterministically and thus obtains identical outputs for identical inputs. 6Img-to-3D also produces temporally consistent images across multiple consecutive timesteps without being explicitly trained to do so. A small change in the pose of the six input cameras results in a small change in the resulting output representation, visualized within the short trajectory in the upper part of \cref{fig:trajectories}. When fed with six images per timestep, our model can also visualize longer scenes for example from a bird's-eye view or a third-person perspective, see the bottom part of \cref{fig:trajectories} long trajectory.

\subsection{Effects of Scaling}\label{sec:effects_scaling}

\begin{figure}[H]
    \centering
    \includegraphics[width=0.5\textwidth]{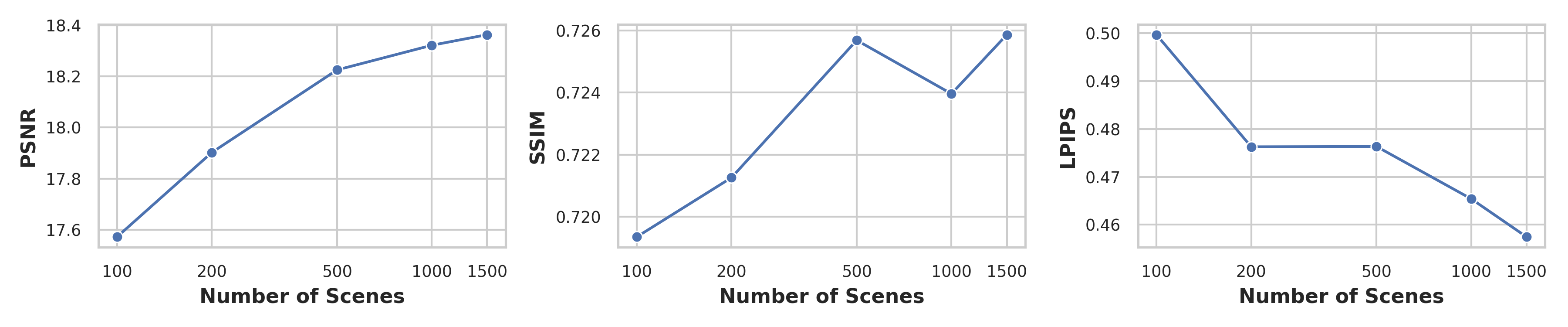}
    \caption{\textbf{Effect of number of training scenes on performance.} Only the number of training scenes was varied, while all other parameters were kept constant.}
    \label{fig:ablation-scenes}
\end{figure}

Since our framework is self-supervised, we investigated whether our pipeline would benefit from a larger dataset. As shown in \cref{fig:ablation-scenes}, the performance scales with the number of scenes. When tested with 100, 200, 500, 1000, and 1500 scenes, we find that a higher number of scenes leads to improved model performance. 

\subsection{Zero-shot nuScenes Performance visualization} \label{sec:zero_shot_nuScenes}

\begin{figure}[h]
    \centering
    \includegraphics[width=1\linewidth]{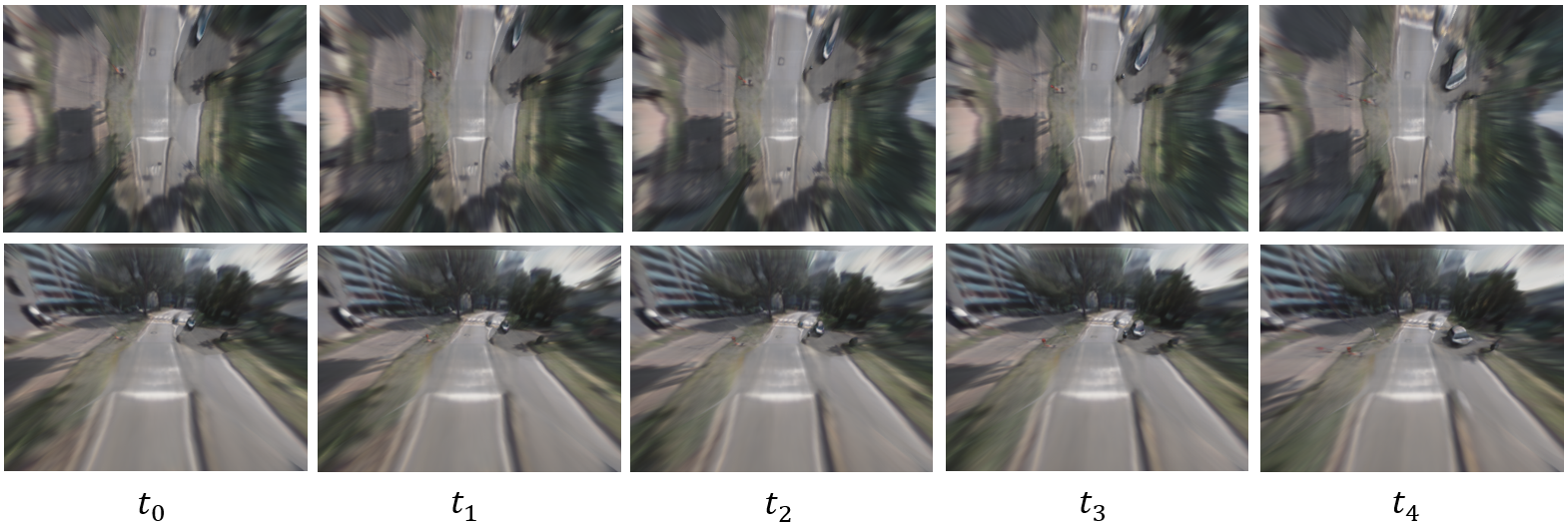}
    \caption{\textbf{nuScenes Short Trajectory.} Consecutive timesteps inferred during zero-shot inference on nuScenes. }
    \label{fig:nuScenes_short_trajectory}
\end{figure}

\begin{figure*}[ht!]
    \centering
    \includegraphics[width=1\linewidth]{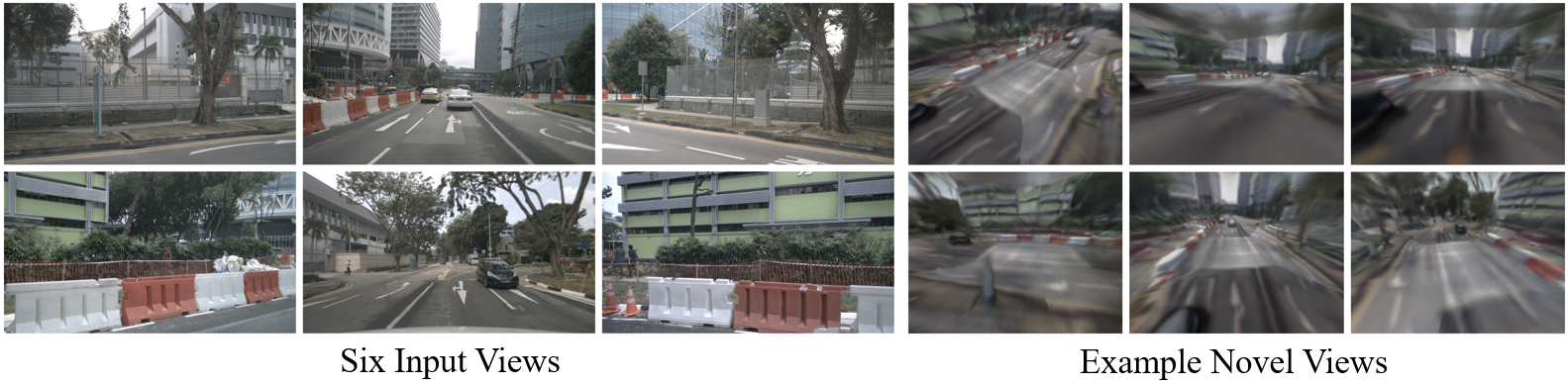}
    \caption{\textbf{6Img-to-3D with nuScenes Data.} By inputting six images from nuScenes into our trained model pipeline, we can generate novel views.   }
    \label{fig:nuScenes_overview}
\end{figure*}

We attempt an early analysis of our 6Img-to-3D applied zero-shot to nuScenes. The results are vizualized in \cref{fig:nuScenes_overview}. Without additional training, the model can display the appearance of nuScenes. Unfortunately, the camera setup within the SEED4D dataset dpes not match the nuScenes data exactly. Camera positions, FOV, and scene style (sunny vs. cloudy weather) differ, making zero-shot transfer difficult. The promising results motivate to extent the model to real-world data. Consistency across multiple timesteps is shown in \cref{fig:nuScenes_short_trajectory}.

%
%
%

\subsection{Failure Cases}

\begin{figure*}[ht!]
    \centering
    \includegraphics[width=1\linewidth]{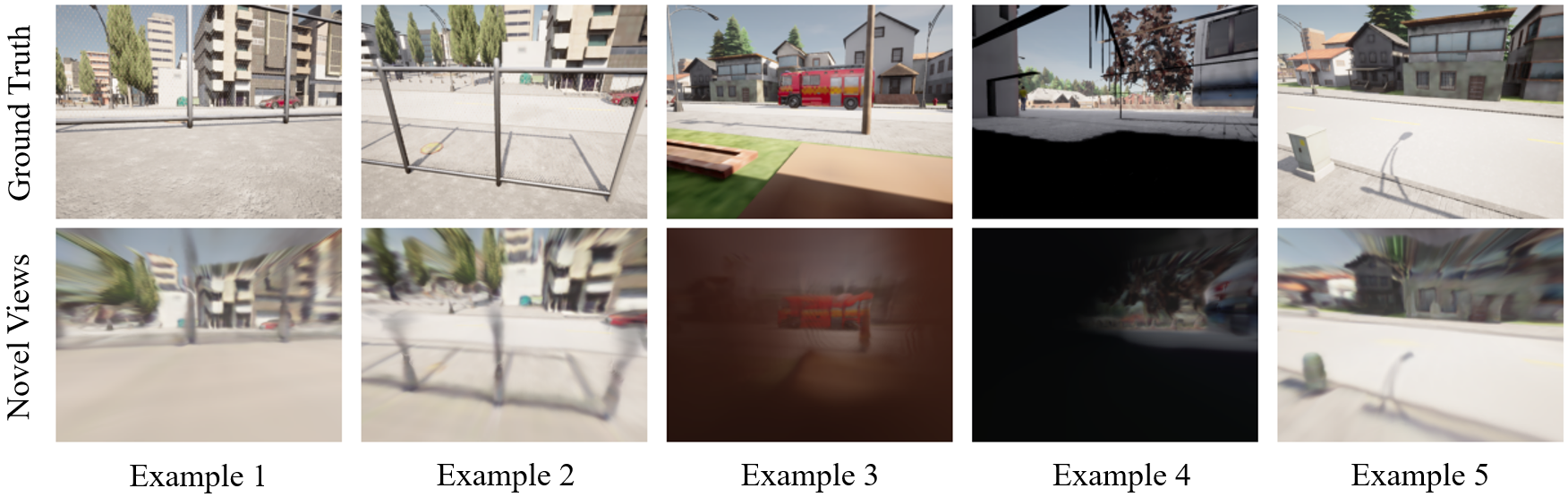}
    \caption{\textbf{Visualization of Failure Cases.} Failure cases can partially be attributed to the CARLA engine, which sometimes struggles to render walls when within a building.}
    \label{fig:failure_cases}
\end{figure*}

In \cref{fig:failure_cases}, failure cases of our method are visualized. Our model struggles with zero-shot reconstructing fine details, as seen in examples one and two. This might be because we only trained the model with an image resolution of $64 \times 48$ pixels. Partially, model failures can be attributed to the dataset itself. Some of the supervising images in the training set are located inside buildings. Within the Carla simulator, walls are sometimes transparent from within the building, which leads the model to pick up on this physically impossible artifact. Those failure cases are shown in examples three and four. Example five shows an instance where the model is unable to infer the structure of the building beyond what is contained in the input images.

\subsection{Extended Input-Output Visualization}

\begin{figure*}
    \centering
    \includegraphics[width=1\linewidth]{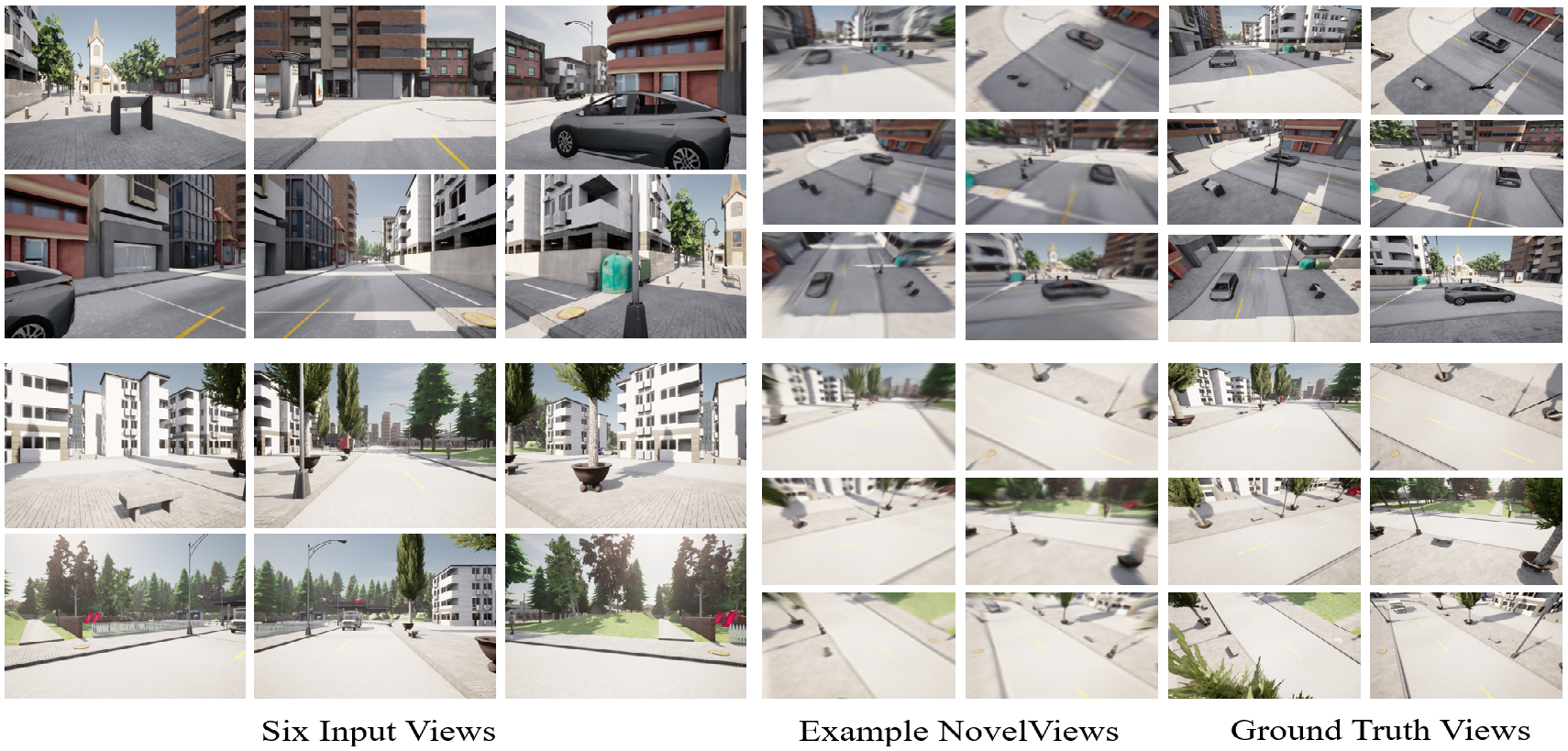}
    \caption{\textbf{Input-Output Examples.} During inference time, 6Img-to-3D takes six surround vehicle RGB images with a small overlap as input and returns a parameterized triplane from which arbitrary novel views can be rendered.}
    \label{fig:input_output_visualization2}
\end{figure*}

To further illustrate the task our model is handling, we visualize several additional input images and example novel views in \cref{fig:input_output_visualization2}. Ego-centric input views are characterized by minimal view overlap, an unbounded line of sight, and frequent occlusions. Both the texture and geometry of objects in a scene vary widely. The exo-centric views represent a diverse mixture of non-ego views. The ego vehicle is not included in the ground truth novel views and, therefore, does also not appear in the rendered novel views.

\subsection{Extended Qualitative Results Comparison} \label{sec:comp_annex}

\begin{figure*}[ht!]
    \centering
      \includegraphics[width=0.9\textwidth]{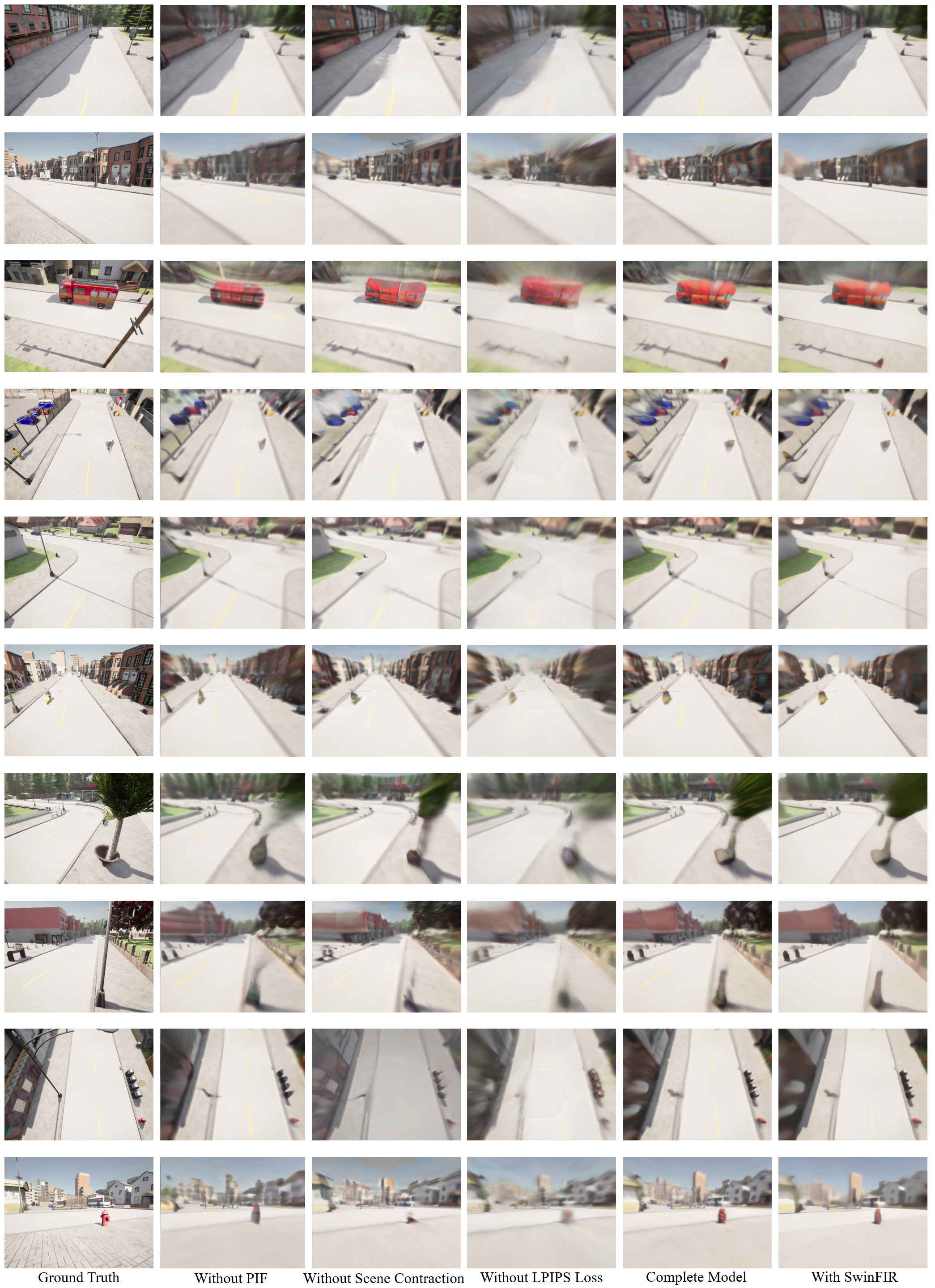}
    \caption{\textbf{Qualitative Ablation Results (I)}.}
    \label{fig:ablation1}
\end{figure*}

\begin{figure*}[ht!]
    \centering
     \includegraphics[width=0.9\textwidth]{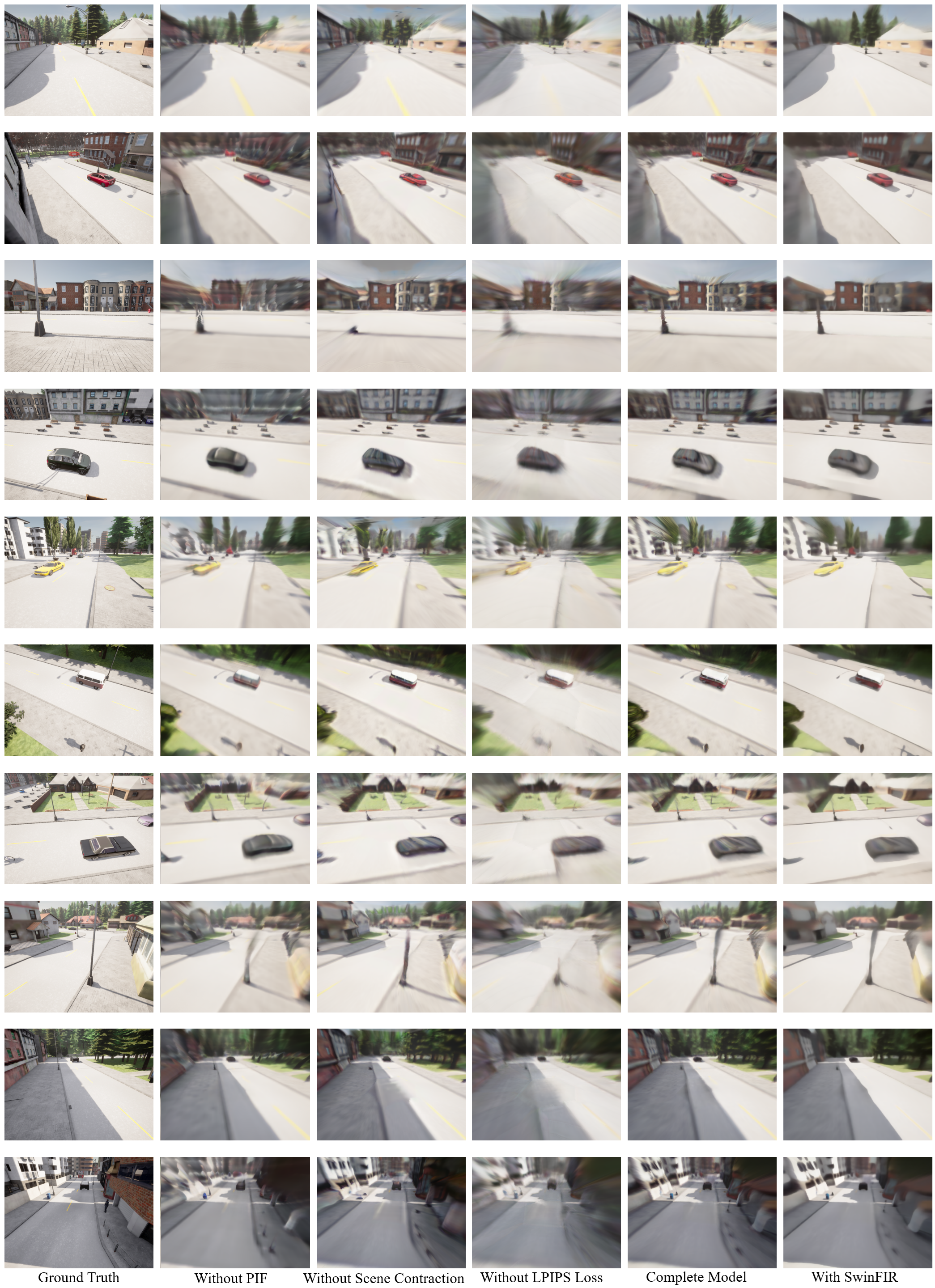}
    \caption{\textbf{Qualitative Ablation Results (II).}}
    \label{fig:ablation2}
\end{figure*}

\begin{figure*}[ht!]
    \centering
      \includegraphics[width=0.9\textwidth]{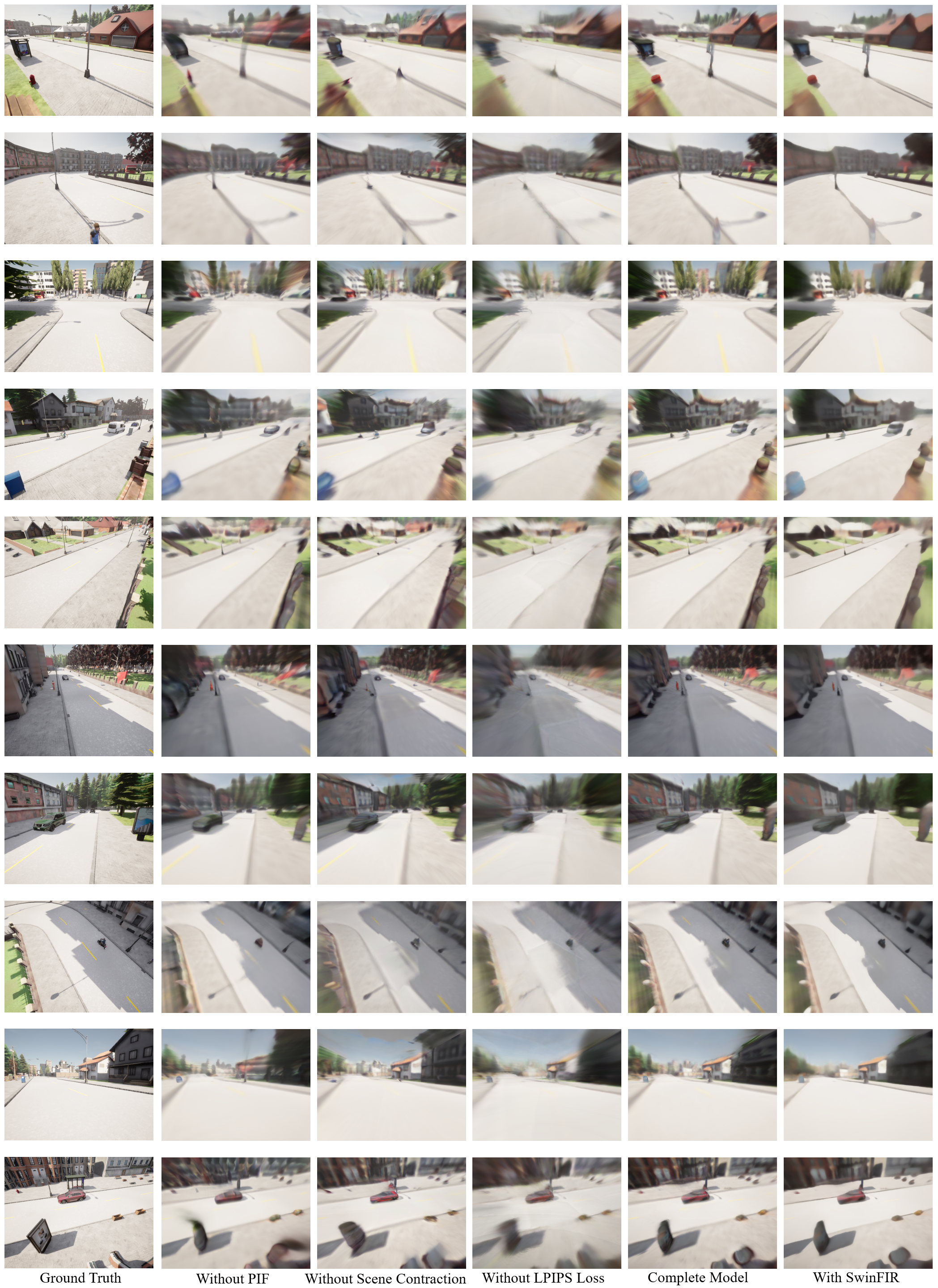}
    \caption{\textbf{Qualitative Ablation Results (III).}}
    \label{fig:ablation3}
\end{figure*}

We extend the visualization from our ablations in \cref{fig:ablation1} to \cref{fig:ablation3} to better understand our model's performance. The results are not cherry-picked. Additionally, an off-the-shelf upscaler, SwinFIR \cite{zhang2023swinfir}, is trained and tested on the output of our 6Img-to-3D model. While slightly improving on all metrics (PSNR 19.188 up from 18.683, SSIM increased to 0.746 from 0.726, and LPIPS decreased to 0.444 from 0.451), a visual analysis shows that the upscaler only outputs a smoothed image instead of adding otherwise missing details. We thus decided not to include the upscaled results in the main analysis.

\subsection{Extended Qualitative Results}

\begin{figure*}
    \centering
    \includegraphics[width=0.85\linewidth]{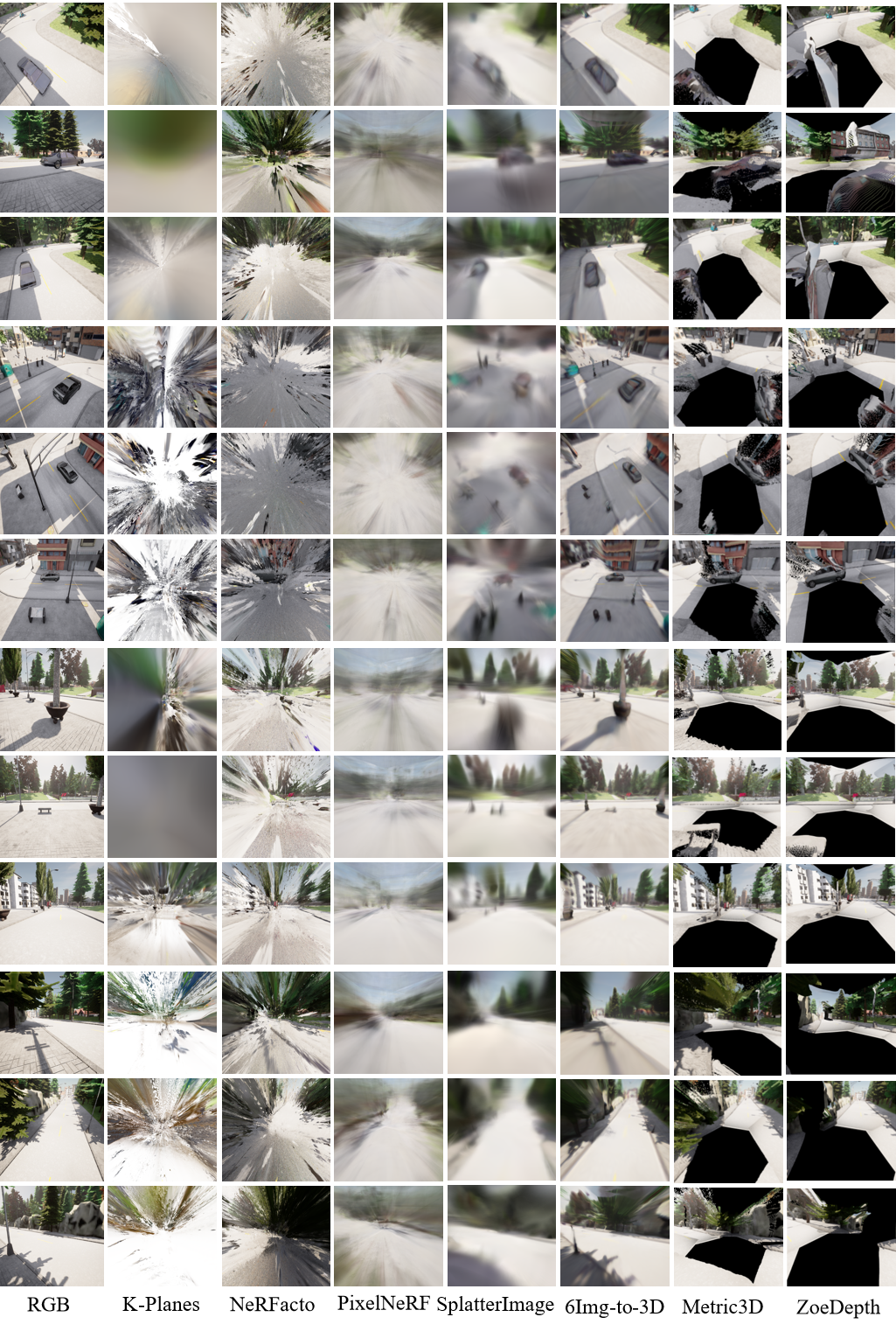}
    \caption{\textbf{Extended Qualitative Results.}}
    \label{fig:extended_qualitative_results_comparison_rgb}
\end{figure*}

\begin{figure*}
    \centering
    \includegraphics[width=0.82\linewidth]{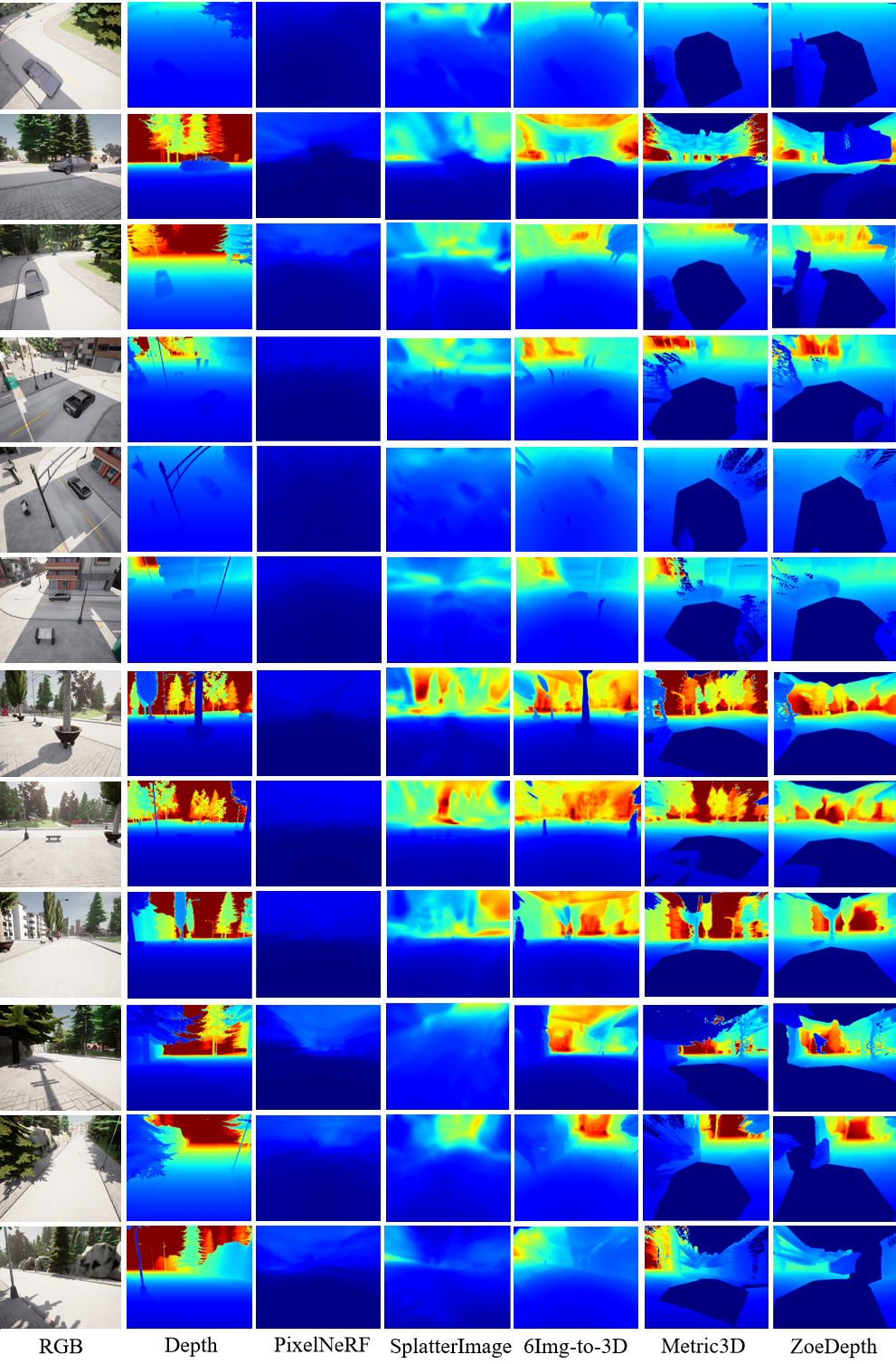}
    \caption{\textbf{Extended Qualitative Results.} Due to their low performance, we do not visualize SplatFacto at the top and SplatFacto K-Planes and NeRFacto at the bottom part.}
    \label{fig:extended_qualitative_results_comparison_depth}
\end{figure*}

This section provides additional qualitative results of reconstructed views and their associated depth maps for various methods. The depictions can be found in \cref{fig:extended_qualitative_results_comparison_rgb} for the rendered images and in \cref{fig:extended_qualitative_results_comparison_depth} for the depth maps.

\subsection{Dataset Visualization} \label{sec:dataset_vis}

Towns differ in buildings (residential and commercial), types of bridges, the number of rivers, the number of traffic lights, and the kind of junctions. All towns contain pedestrians. For further details about the dataset, please see the dataset publication \cite{SEED4D_unpublished} or the additional visualization we provide in \cref{sec:dataset_vis}.

\begin{figure*}
    \centering
    \includegraphics[width=1\linewidth]{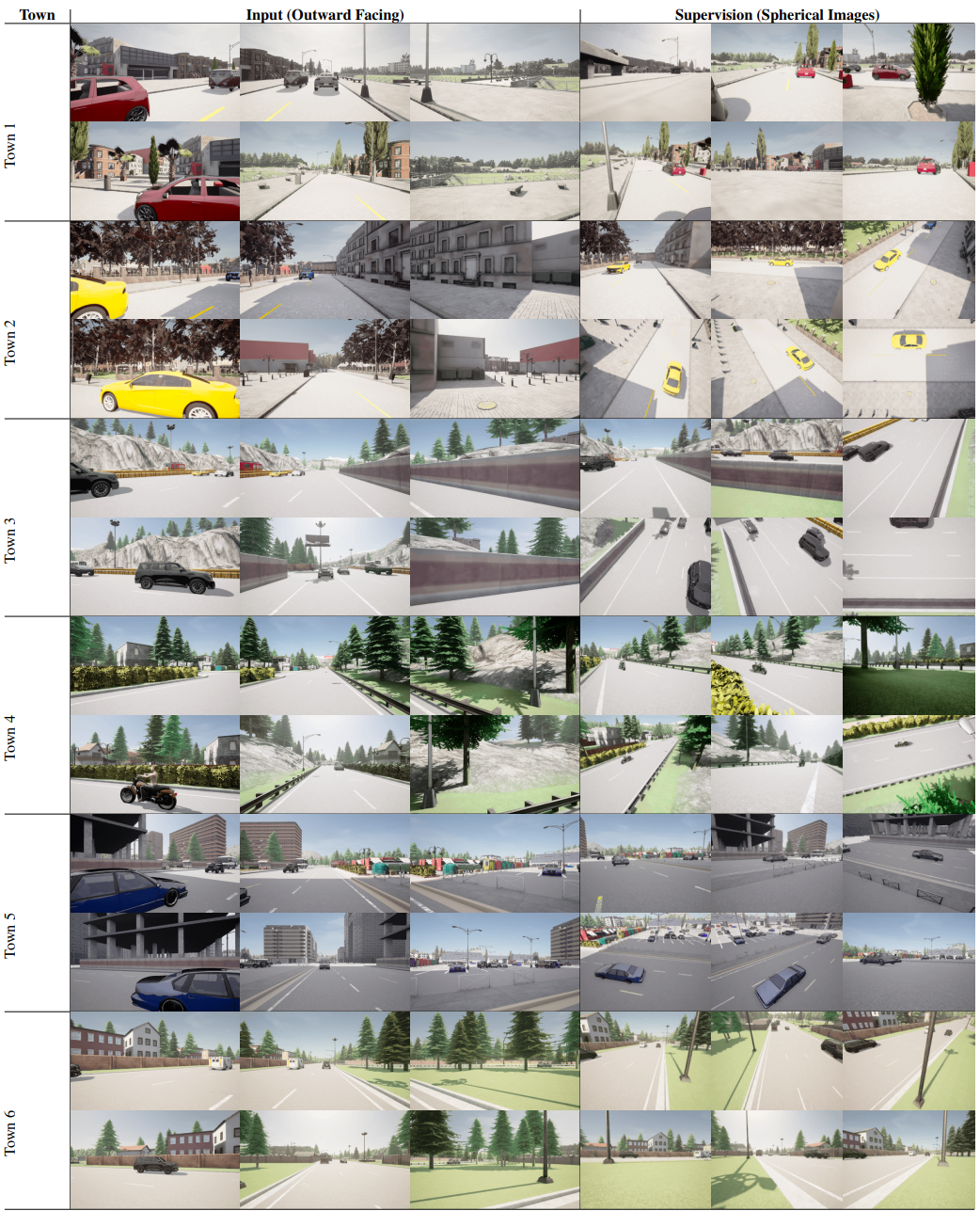}
    \caption{\textbf{Samples from the SEED4D Dataset.} Ego- and Exo-centric images for a number of towns.}
    \label{tab:datasetsample}
\end{figure*}

\end{document}